\documentclass[lettersize,journal]{IEEEtran}
\usepackage{xcolor}
\usepackage{amsmath,amssymb,amsfonts}
\usepackage{bm}
\usepackage{mathtools}
\usepackage{algorithmic}
\usepackage[ruled,linesnumbered]{algorithm2e}
\usepackage{array}
\usepackage[caption=false,font=normalsize,labelfont=sf,textfont=sf]{subfig}
\usepackage{textcomp}
\usepackage{stfloats}
\usepackage{url}
\usepackage{verbatim}
\usepackage{graphicx}
\usepackage{cite}
\usepackage{multirow}
\usepackage{soul}

\IEEEaftertitletext{\vspace{-2\baselineskip}}

\begin{document}

\title{{GNN-based} Auto-Encoder for Short Linear Block Codes: {A DRL Approach}}

\author{Kou Tian,~\IEEEmembership{Student Member,~IEEE,}
        Chentao Yue,~\IEEEmembership{Member,~IEEE,}
        Changyang She,~\IEEEmembership{Senior Member,~IEEE,}\\
        Yonghui Li,~\IEEEmembership{Fellow,~IEEE,}
        and Branka Vucetic,~\IEEEmembership{Life Fellow,~IEEE}
      

\thanks{{K. Tian, C. Yue, Y. Li, and B. Vucetic are with the School of Electrical and Computer Engineering, The University of Sydney, Sydney, Australia.
C. She is with School of Electronic and Information Engineering, Harbin Institute of Technology (Shenzhen), Shenzhen, China. Part of this work was done when he was with the School of Electrical and Computer Engineering, The University of Sydney, Sydney, Australia.} 
{E-mails:\{kou.tian, chentao.yue, yonghui.li, branka.vucetic\}@sydney.edu.au, shechangyang@gmail.com}}
\thanks{This paper was presented in part at IEEE International Conference on Communications (ICC), Rome, Italy, June 2023. This work has been supported by the SmartSat CRC, whose activities are funded by the Australian Government’s CRC Program.}
}

\markboth{Journal of \LaTeX\ Class Files,~Vol.~14, No.~8, August~2021}%
{Shell \MakeLowercase{\textit{et al.}}: A Sample Article Using IEEEtran.cls for IEEE Journals}


\maketitle

\begin{abstract}
This paper presents a novel auto-encoder based end-to-end channel encoding and decoding. It integrates deep reinforcement learning (DRL) and graph neural networks (GNN) in code design by modeling the generation of {code} parity-check matrices as a Markov Decision Process (MDP), {to optimize} key {coding} performance metrics such as error-rates and code algebraic properties. 
An edge-weighted GNN (EW-GNN) decoder is {proposed}, which operates on the Tanner graph with an iterative message-passing structure. 
{Once trained on a single {linear block} code, the EW-GNN decoder can be directly used {to decode other {linear block} codes of different code lengths and code rates}.}
{An iterative joint training of the DRL-based code designer and the EW-GNN decoder is performed to optimize the end-end encoding and decoding process.}
Simulation results show the proposed auto-encoder significantly surpasses several traditional coding schemes at short block lengths, {including low-density parity-check (LDPC) codes with the belief propagation (BP) decoding {and} the maximum-likelihood decoding (MLD), and BCH with BP decoding}, offering superior error-correction capabilities while maintaining low decoding complexity.
\end{abstract}
 
\vspace{-0.6em}
\section{Introduction}
\vspace{-0.2em}
The {increasing} demand for ultra-reliable low-latency communications (URLLC) in next{-}generation communication networks {requires} {low-latency} physical-layer designs {--} particularly channel coding schemes.
Short codes with strong error-correction capability  {are essential to} satisfy the stringent latency and reliability requirements of URLLC in 5G and future 6G networks \cite{shirvanimoghaddam2018short, yue2022efficient}.
To meet these requirements, the channel coding system must demonstrate superior error-rate performance and low decoding complexity.
{This necessitates} {both} a well-designed short channel code at the transmitter side and {an efficient} decoder at the receiver side.

Several short linear block codes, including Bose-Chaudhuri-Hocquenghem (BCH) codes, LDPC codes {and Polar codes} are promising candidates for URLLC due to their superior performance in the short block length {regime} \cite{shirvanimoghaddam2018short}.
Although long LDPC codes with the BP decoder have been adopted in the data channel of 5G new radio \cite{3gpp_nrcoding} due to their near-capacity {performance} and reasonable decoding complexity, their short variants suffer from a significant {performance} gap {compared} to the finite block length bound \cite{polyanskiy2010channel}.
The suboptimal {performance} of short LDPC codes primarily {arises from the insufficient sparsity in the parity-check matrices, resulting in {inherent} limitation{s} of the BP decoder, which {exhibits} a notable {block-error-rate (BLER)} performance degradation compared to MLD \cite{shirvanimoghaddam2018short}.}

The conventional design of coding schemes involves {dealing with} a trade-off between {BLER} performance and decoding complexity. 
{Such designs usually rely on the probabilistic and algebraic analyses.
Due to the complexity of {these} analyses, the encoder and decoder are designed and optimized separately.}
{Codes with high error-correction capabilities} require complex decoding algorithms. 
For example, short BCH codes {{are among} the best-known codes with strong error-correction performance, while their optimal decoding} is {highly} computationally expensive {\cite{shirvanimoghaddam2018short}}. 
{On the other hand,} {LDPC paired with BP} exhibits {a significant BLER gap to the capacity bound}, particularly {at} short block lengths \cite{shirvanimoghaddam2018short}. 

{Recently, t}here is a growing interest in leveraging artificial intelligence (AI) for wireless communication and channel coding \cite{DLforWN_survey, DLforWC_2020}. 
AI, particularly deep learning (DL), can potentially surpass {analytical methods} {in the design of communication systems} by {learning the channel statistics, extracting features from wireless networks and exploring the optimal communication algorithms.} 
AI-driven approaches have shown significant advantages in {wireless system design and optimization}, including {signal detection \cite{samuel2017deepmimo} and channel resource allocation\cite{sun2017resouce}}. 
AI has also been integrated into {code designs} \cite{LEARNcodes, kim2020deepcodes, ebada2019deeppolar, huang2020AIcoding}, {decoder designs} \cite{deepcoding, nbp2016, nbp2018, hyperbp, buchberger2020pruning, Nachmani2022activation, cammerer2022gnndec, ewgnn}, and the joint optimization of {encoding and decoding} via end-to-end training \cite{nbp_ae2022, into_dl_phy, conv_ae, jiang2019turboae, 1b_ae}. 
The architecture that implements both the encoder and decoder with AI models is referred to as ‘‘auto-encoder'' channel coding.

\vspace{-0.6em}
\subsection{Related Works}
\vspace{-0.2em}
\subsubsection{DL for Code Design}
The powerful capabilities of DL in data processing and feature learning offer a novel approach for constructing channel codes that can {potentially} outperform those designed by using human ingenuity \cite{LEARNcodes, kim2020deepcodes, ebada2019deeppolar, huang2020AIcoding}. 
{In \cite{LEARNcodes, kim2020deepcodes}, the authors used recurrent neural networks (RNNs) to construct low-latency and feedback codes, respectively, demonstrating performance gains.}
{Authors in} {\cite{ebada2019deeppolar} proposed DL-based polar codes tailored to BP decoding, by treating the frozen positions of a polar code as trainable weights of a neural network.}
{These works utilized existing code structures {to ensure a manageable} training complexity. 
{Generally, }learning to design linear block codes {is challenging and {faces several major challenges}:}}
1) \textit{Curse of dimensionality} \cite{donoho2000high_dim}: As the {message length} $k$ increases, the volume of the search space (i.e., the number of possible messages and codewords) increases exponentially as $2^k$. {Thus,} the required number of training data {becomes impractically large for supervised learning}. Furthermore, a large $k$ also {results in} high-dimensional input data, which leads to increased computational complexity.
2) \textit{Gradient calculation}: {Code matrices, i.e., t}he generator matrix and the parity-check matrix{,} are binary, which complicates the gradient calculation in the DL training process.
3) \textit{Matrix validity}: {Code matrices must be of full rank to ensure the code linearity, which} introduces {an extra} constraint on the learning process.
{In \cite{huang2020AIcoding}, reinforcement learning (RL) was used to construct the binary code generator matrix in {a} {systematic} form. However, the designed codes exhibit limited error-rate performance.}

\subsubsection{DL for Channel Decoding}
Recent works have achieved notable improvements in {using} DL-aided decoders {to decode} traditional analytically designed codes.
{D}eep neural networks based on the structure of {the} existing decoding algorithms, such as BP{, have been extensively explored}.
In \cite{nbp2016} and \cite{nbp2018}, the authors proposed a neural BP (NBP) {decoder}, by unrolling the iterative structure of BP to a non-fully connected network with trainable weights. NBP outperforms the conventional BP in terms of {BLER when decoding} short BCH codes. 
The NBP decoder {was} further improved in \cite{hyperbp} {by integrating it with a hyper-graph network, and} {it} {was also} {enhanced} by {the} pruning-based NBP {in} \cite{buchberger2020pruning}, which removes less-weighted neurons in the NBP network to obtain different parity-check matrices in each iteration.
In addition, {the loss functions for training NBP {were refined} in \cite{Nachmani2022activation}.} 
Motivated by the {graph-based representation of codes using} Tanner graphs, \cite{cammerer2022gnndec} proposed a GNN decoder for linear block codes, {which} demonstrates competitive {BLER} performance for LDPC and BCH codes.
Although aforementioned DL-aided {decoders} {have} made significantly progress, {they have several limitations.} 
{Speficially, }1) {their} BLER performance gap to the optimal MLD {remains significant};
2) {they are not scalable to code length and code rate, requiring re-training when either changes;} 
{and} 3) {their performance} highly depends on the hyper-parameters of neural networks, which are obtained by trail-and-error.

\subsubsection{Auto-Encoder}
{The traditional} coding system integrates individually optimized {encoder and decoder}, but this approach does not necessarily yield a globally optimal solution.
Recently, DL has been used for the global optimization of {``auto-encoder'' }channel coding systems with jointly designed encoder and decoder \cite{nbp_ae2022, into_dl_phy, conv_ae, jiang2019turboae, 1b_ae}.
However, due to the high complexity of training end-to-end models and the {aforementioned} dimensionality problem, the application of auto-encoders is often limited to very small code sizes. 
For example, the auto-encoder proposed in \cite{into_dl_phy} can only be utilized for codeword lengths up to 8. 
One solution is to design structured auto-encoder models {aligning with specific code structures}, such as convolutional {auto-encoder (ConvAE)} \cite{conv_ae} and Turbo {auto-encoder (TurboAE)} \cite{jiang2019turboae}, rather than using a direct black-box model as in \cite{into_dl_phy}. 
{ConvAE uses convolutional neural networks (CNN) for block codes, and TurboAE is a turbo {code-like} {variant} of ConvAE, utilizing multiple CNNs for interleaved encoding and iterative decoding. However, these methods {cannot learn good codes at}  moderate-to-long {lengths}, offering only slight improvements over {analytically} designed codes.}
In \cite{nbp_ae2022}, the authors proposed a neural belief propagation auto-encoder (NBP-AE), {which uses {trainable binary weights that represent the parity-check matrix} to link the encoder and decoder to enable end-to-end training}.
However, NBP-AE still has a significant BLER performance gap to the carefully designed encoder-decoder pairs, including BCH codes with its near MLD and irregular LDPC codes with BP.

\vspace{-0.7em}
\subsection{Contributions}
\vspace{-0.2em}
This paper proposes a GNN-DRL auto-encoder. 
First, we develop a DRL-based code designer that can learn better codes with superior error-control capabilities compared to conventional codes. 
Then, we develop an EW-GNN decoder.
Finally, the DRL-based code designer and the EW-GNN decoder {are jointly optimized and integrated into} an auto-encoder channel coding framework.
The main contributions of the paper are summarized as follows:

\begin{itemize}
    \item \textit{DRL-based code designer}: 
    We introduce a DRL-based code designer that leverages GNNs to construct high-performance parity-check matrices for linear block codes. 
    {This approach formulates the process of generating matrices as a MDP, where a GNN-based agent directly operates within the parity-check matrix by treating it as a {lattice} graph.}
    {Our} DRL-based method supports active exploration{s} of the code space beyond given data samples, {and {the reward functions are} designed to optimize critical code performance metrics, including girth, decoding bit error rate (BER), and the minimum Hamming distance (MHD)}.
    \item \textit{Scalable GNN decoder}: 
    We develop a novel EW-GNN decoder for short linear block codes. 
    To simplify the selection of hyper-parameters, EW-GNN 
    {applies the \textit{algorithmic alignment} framework \cite{algorithm_aligen} by designing a unique GNN that precisely matches the traditional BP decoding method and operates over Tanner graphs.}
    Specifically, each variable node in EW-GNN is assigned a hidden embedding that is iteratively updated based on edge messages on the Tanner graph.
    Then, each edge message is assigned a ``weight'' {determined by a fully-connected feed-forward neural network (FNN).
    The FNNs located on all edges share the same trainable parameters.} 
    By reducing the weights of unreliable edge messages generated from small cycles {of Tanner graphs}, EW-GNN {significantly }improves the {BER compared to BP}.
    {Due to the intrinsic scalability properties of GNNs, }the number of trainable parameters of EW-GNN does not change with the code length. 
    {Moreover}, after {the supervised training process} with a given linear block code, {it} can be applied to other {linear block} codes with different rates and lengths.
    
    \item \textit{Jointly-designed auto-encoder}: 
    We present an auto-encoder framework that combines the proposed DRL-based code designer with the EW-GNN decoder, enabling the joint optimization of {codes and their decoders}. 
    An iterative training scheme is applied to the proposed auto-encoder model. 
    {First,} the DRL-based code designer {designs the} optimal code, using a reward function {that minimizes the BER with} {an initial} decoder, {e.g., BP}.
    {Next,} the EW-GNN decoder {is trained for the designed code}.
    {In the next training iteration, a new code will be updated based on the BER reward from the EW-GNN decoder, and the decoder will be fine-tuned again for the new code. This training process will continue for several iterations.}
    
    \item 
    {We conducted extensive simulations to compare our code designer, decoder, and auto-encoder to conventional methods.} 
    The DRL-based code designer can discover {better codes than} LDPC codes under {both} BP and MLD {decoders}. Under MLD, the learned code {has} $0.83~\mathrm{dB}$ {more} coding gain {than} the $(32,16)$ LDPC code {on additive white Gaussian noise (AWGN) channels}.
    {Moreover,} EW-GNN {decoder} outperforms BP and NBP in terms of BER, even when the code length in the training stage differs from that in the testing stage. For the $(63, 51)$ BCH code, EW-GNN increases the coding gain by $1.2~\mathrm{dB}$ and $0.62~\mathrm{dB}$ compared to BP and NBP decoders, respectively.
    The proposed auto-encoder framework {provides} superior error-rate performance compared to multiple conventional encoder-decoder pairs {and NBP-AE \cite{nbp_ae2022}} at short block lengths. {For example}, {our auto-encoder} outperforms {the $(32,16)$} LDPC decoded {with} MLD {and BP} by a coding gain of $0.63~\mathrm{dB}$ {and $1~\mathrm{dB}$, respectively}. 
\end{itemize}

The rest of this paper is organized as follows.
Section \ref{sec::preliminaries} provides the preliminaries. 
In Section \ref{sec::DRL_code_designer}, we introduce the DRL-based code designer. 
Section \ref{sec::EW-GNN} presents the EW-GNN decoder, outlining its structure and training algorithm.
Section \ref{sec::auto_encoder} presents the iterative training process of the proposed auto-encoder framework. 
In Section \ref{Sec::experiments}, we provide simulation results.
Finally, Section \ref{sec::conclusion} concludes the paper.


\vspace{-0.7em}
\section{Preliminaries} \label{sec::preliminaries}
\vspace{-0.3em}
\subsection{Channel Coding with Linear Block Codes} \label{sec::lbc}
\vspace{-0.3em}
We consider a binary linear block code {denoted by} $\mathcal{C}(n,k)${, where} a $k$-bit binary message sequence is encoded {into} a codeword of length $n$ with $n > k$.

\subsubsection{Matrix Representation} 
For {$\mathcal{C}(n,k)$}, the encoding process can be succinctly represented by 
\begin{equation} \label{equ::enc}
\mathbf{c} = \mathbf{b}\mathbf{G},
\end{equation}
where $\mathbf{b} \in \mathbb{F}_2^k$ is the raw message, $\mathbf{c} \in \mathbb{F}_2^n$ is the codeword, and $\mathbf{G} \in \mathbb{F}_2^{k \times n}$ is the {code} generator matrix.
{Eq. \eqref{equ::enc}} is performed over the {binary} field $\mathbb{F}_2$.
We denote the MHD of $\mathcal{C}(n,k)$ as $d_{\min}$.

The parity-check matrix of $\mathcal{C}(n,k)$ is denoted as $\mathbf{H} \in \mathbb{F}_2^{m \times n}$, where $m=n-k$. 
A parity-check matrix $\mathbf{H}$ can be transformed into the {systematic} form as $\mathbf{H}_{\mathrm{std}} = (\mathbf{A}\mid\mathbf{I}_m)$ through \textit{Gaussian elimination}, where $\mathbf{A}$ {$\in \mathbb{F}_2^{m\times k}$} and $\mathbf{I}_m$ is an identity matrix of size $m$. 
The associated {systematic} generator matrix can be {accordingly} obtained as \(\mathbf{G}_{\mathrm{std}} = (\mathbf{I}_k\mid\mathbf{A}^\intercal)\). 
{Thus, $\mathbf{G}$ and $\mathbf{H}$ are both of full-rank, and either of them can fully define $\mathcal{C}(n,k)$.}

\subsubsection{Graphical Representation}
\label{Sec::Preliminaries::TG}

\begin{figure}[tb]
\centering
\includegraphics[width=0.48\textwidth]{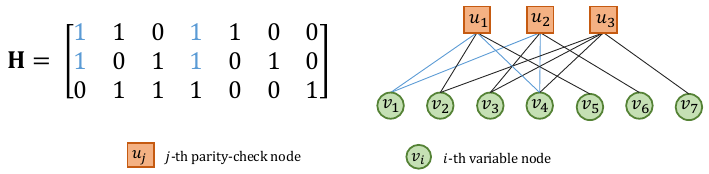}
\vspace{-0.4em}
\caption{A parity-check matrix in systematic form and its corresponding Tanner graph. The path marked by blue lines represents a 4-length cycle.}
\vspace{-1.5em}
\label{fig::tannergraph}
\end{figure}

As illustrated in Fig. \ref{fig::tannergraph}, {the} Tanner graph {of $\mathcal{C}(n,k)$} consists of two sets of nodes: $n$ variable nodes corresponding to codeword bits and $m$ check nodes corresponding to {parity-check bits}. 
The $i$-th variable node $v_i$ is connected to the $j$-th check node $u_j$ if the $(j,i)$-th element of $\mathbf{H}$ is one. 
In this case, H can be interpreted as the adjacency matrix of a bipartite graph $ \mathcal{G}=(\mathcal{V}, \mathcal{U}, \mathcal{E})$ with two disjoint node sets $\mathcal{V} = \{v_1,...,v_{n}\}$ and $\mathcal{U} = \{u_1,...,u_{m}\}$, and an edge set defined as $\mathcal{E} = \{e_{i,j} = (v_i, u_j): v_i\in\mathcal{V},u_j\in\mathcal{U}, H_{j,i}=1\} $. We denote the neighborhood of a variable node $v_i$, i.e., all check nodes connected to $v_i$, as  $\mathcal{M}(v_i) = \{u_j \in \mathcal{U} : e_{i,j} \!\in \!\mathcal{E}\}$. Similarly, $\mathcal{M}(u_j) = \{v_i \in \mathcal{V} : e_{i,j} \!\in\! \mathcal{E}\}$ represents the set of neighbor variable nodes of a check node $u_j$. 

In the Tanner graph $\mathcal{G}$, a \textit{cycle} of length $\lambda$ (i.e., {a} $\lambda$-cycle) is a closed path that starts and ends at the same node, traversing $\lambda$ edges in $\mathcal{E}$ without repetition. The \textit{girth} {$g$} of $\mathcal{G}$ is the length of the shortest cycle. 
Since $\mathcal{G}$ is bipartite, we have $g \geq 4$. 
The girth {affects} the {decoding} performance {when using BP}\cite{heuristicLDPC}.

\subsubsection{Belief Propagation Decoding}
BP is a soft-decision decoding algorithm \cite{LLRbp}. We denote the channel log-likelihood ratio (LLR) sequence of codeword $\mathbf{c}$ as $\bm{\ell} \in \mathbb{R}^n$. For the binary phase-shift keying (BPSK) modulation over an AWGN channel, $\ell_i$ can be obtained by 
\begin{equation} \label{equ::llr}
\ell_i = \ln{\frac{\mathrm{Pr}(c_i = 0\mid y_i)}{\mathrm{Pr}(c_i = 1 \mid y_i)}} = \frac{2y_i}{\sigma_n^2}.
\end{equation}
{With $\bm{\ell}$}, BP performs iterative message-passing {decoding}. Specifically, at the {$t$-th iteration}, it performs
\begin{equation} \label{equ::u_to_v}
    \mu_{u_j \to v_i}^{(t)} = 2\tanh^{-1}\left(\prod_{v \in \mathcal{M}(u_j)\setminus v_i} \tanh \left(\frac{\mu_{v \to u_j}^{(t-1)}}{2} \right) \right),
\end{equation}
for each $u_j \in \mathcal{U}$ and $v_i \in \mathcal{M}(u_j)$, and  
\begin{equation}  \label{equ::v_to_u}
    \mu_{v_i \to u_j}^{(t)}= \ell_i + \sum_{u \in \mathcal{M}(v_i)\setminus u_j}\mu_{u \to v_i}^{(t)},
\end{equation}
for each $v_i \in \mathcal{V}$ and $u_j \in \mathcal{M}(v_i)$.
BP iteratively performs \eqref{equ::u_to_v} and \eqref{equ::v_to_u}, until $\mu_{u_j \to v_i}^{(t)}$ and $\mu_{v_i \to u_j}^{(t)}$ converge, or the allowed maximum number of iterations is reached. 

The {BER} of BP is strongly related to the {girth $g$}. 
{S}mall cycles introduce correlations between {messages passed across decoding iterations,} thereby preventing the \textit{a posteriori} LLR being exact {\cite{fgandspa}}. 
{F}or BCH codes and short LDPC codes, BP has inferior performance due to {their} small girth{s} \cite{multi-baseBP, shirvanimoghaddam2018short}. 

\vspace{-0.9em}
\subsection{Learning Approaches}
\vspace{-0.2em}
\subsubsection{Reinforcement Learning} RL is a trial-and-error approach to learn the optimal policy for sequential decision-making in dynamic environments{. In RL, }the decision-maker, {known as the agent}, performs actions in an environment and refines its behavior based on feedback received as rewards or penalties. A basic RL framework can be described by the MDP {with components: State $\mathbf{S}^t$, Action $\mathbf{A}^t$, Policy $\mu$, and Reward $r$. Precisely, policy determines actions in given states, i.e., $\mathbf{A}^t = \mu (\mathbf{S}^t)$. The reward evaluates action quality to guide the agent's behavior, which is given by $r^t = R(\mathbf{S}^t, \mathbf{A}^t, \mathbf{S}^{t+1})$ with a reward function $R$.} In the agent-environment interaction loop, the agent takes an action $\mathbf{A}^t$ based on the current {policy} {$\mu$ and $\mathbf{S}^{t}$}, then the {system transits} with a new state $\mathbf{S}^{t+1}$ and {receives} a reward $r^t$. 
The goal of RL is to learn an optimal policy {that can} maximize the {long-term} reward, which is estimated by the action-value (Q-value) function in most RL algorithms. 

In this work, we adopt the deep deterministic policy gradient (DDPG) algorithm \cite{lillicrap2015ddpg}, {which is} an actor-critic DRL method {combining} the strengths of the policy optimization method \cite{silver2014dpg} and the Q-learning method \cite{DQN}.
{I}t applies two deep neural networks to simultaneously learn a deterministic policy and an action-value function.
The actor network determines the optimal actions, and the critic network estimates the Q-value, i.e., {the long-term reward.}
{DDPG has more stable and faster convergence {than traditional RL} by combining policy gradient for optimizing policies and Q-learning for value estimation}.

\subsubsection{Graph Neural Network}
GNN is developed for processing graph structured data.
It learns the low-dimensional representation of graphs, i.e., the network \textit{embedding}, by iteratively propagating and aggregating the node/edge information over the graph \cite{gnnsurvey}. Embeddings capture the graph topology, and have been used for graph analysis tasks, e.g., node/graph classification and link prediction \cite{gnnsurvey}. 

Message passing neural networks (MPNN)\cite{mpnn} {have been successfully implemented in wireless communications \cite{GNNforWC, cammerer2022gnndec}}. 
For a input graph, MPNN generates node embeddings {through two neural network-based processes: the message calculation process and the embedding updating process}. 
For each edge in the graph, a ``message'' {is computed} based on the node embeddings of its endpoints. Then, for each node, {MPNN} updates the embedding value by aggregating messages from its neighbors.  
{Due to the graph-based nature of GNNs, the message calculation and the embedding updating can be applied to process the information flow on the Tanner graph, as demonstrated in the EW-GNN decoder ({see} Section \ref{sec::EW-GNN_structure}). 
In the DRL-based code designer, the message calculation and the embedding updating are designed to operate on the lattice graph related to the parity-check matrix of a linear block code ({see} Section \ref{sec::enc::GNN}).}

\vspace{-0.6em}
\section{Code Design with {DRL}}
\vspace{-0.2em}
\label{sec::DRL_code_designer}
{The} enumeration algorithm{s} to find {short} codes {with the largest MHD} \cite{Grassl:codetables} involve prohibitively high computational complexity, and the resulting codes often lack efficient decoding algorithms.
To {efficiently} {construct codes with superior error-correction performance and low decoding complexity}, we {propose} a {DRL-based} code designer, {which leverages} DDPG to {produce {code} parity-check matrices}.  

\vspace{-0.6em}
\subsection{System Model}
\vspace{-0.2em}

\begin{figure}[tb]
\centering
\includegraphics[width=0.4\textwidth]{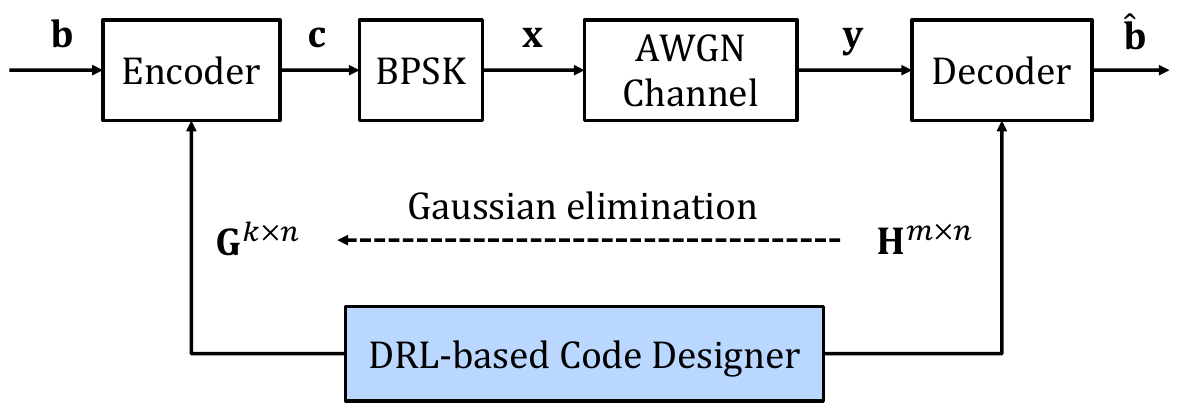}
\vspace{-0.4em}
\caption{System model. }
\vspace{-1.4em}
\label{fig::sysmodel}
\end{figure}

As demonstrated in Fig. \ref{fig::sysmodel}, we consider a binary code $\mathcal{C}(n,k)$ defined by a parity-check matrix $\mathbf{H} \in \mathbb{F}_2^{m \times n}$, which is {generated} by the DRL-based code designer.
Then, the corresponding generator matrix $\mathbf{G} \in \mathbb{F}_2^{k \times n}$ can be obtained by \textit{Gaussian elimination}.
{A binary message $\mathbf{b}$ is then encoded to a codeword $\mathbf{c}$ following \eqref{equ::enc}}. 
We assume that $\mathbf{c}$ is transmitted over the AWGN channel with the BPSK modulation. Thus, {the modulated symbol} $\mathbf{x}$ is given by $\mathbf{x} = 1 - 2{\mathbf{c}}\in \{-1, 1\}^n$, and the noisy received signal $\mathbf{y}$ is obtained as $\mathbf{y} = \mathbf{x} + \mathbf{z}$, where $\mathbf{z}$ is the AWGN vector with {each} $z_i \sim \mathcal{N}(0, \sigma_n^2) $. {The signal-to-noise ratio (SNR) is defined as $\mathrm{SNR} = 1/\sigma_n^2$.}

At the receiver side, a decoder processes $\mathbf{y}$ and outputs an estimate $\hat{\mathbf{c}}$ of $\mathbf{c}$. 
The estimated message $\hat{\mathbf{b}}$ {is} the first $k$ bits of $\hat{\mathbf{c}}$ when the generator matrix is systematic.
A {de}coding error occurs when $\hat{\mathbf{b}} \neq \mathbf{b}$. 
{T}he BER {performance of the code defined by $\mathbf{H}$} {is} calculated as:
\begin{equation}
    {\epsilon_b}(\mathbf{H}) = \mathbb{E}\left[\frac{1}{k} \sum_{i=1}^{k}\mathbf{1}_{[b_i \neq \hat{b}_i]}\right],
\end{equation}
{where $\mathbf{1}_{[b_i \neq \hat{b}_i]} = 1$ if $b_i \neq \hat{b}_i$, and $\mathbf{1}_{[b_i \neq \hat{b}_i]} = 0$, otherwise. $\mathbb{E}[\cdot]$ is the expectation taken over random AWGN noise vectors.}

\vspace{-0.8em}
\subsection{Problem Formulation}
\vspace{-0.2em}
\label{Sec::problem_formulation}
The process of constructing parity-check {matrices} can be treated as a {Tanner} graph structure generation task.
{Intuitively,} {one} can use a DRL agent to modify one edge at each time step, until all possible edges are updated. 
However, for the Tanner graph of $\mathcal{C}(n,k)$, the number of steps in each episode {can be} {as high as} $m \times n${, which leads to high training complexity.} 
In addition, {{a reward is} only available after the graph is fully generated}. 
{This issue is known as the \textit{sparse reward} because the agent receives feedback only at the end of the entire episode, making it challenging to identify the impact of each individual action.}

\begin{figure}[tb]
\centering
\includegraphics[width=0.48\textwidth]{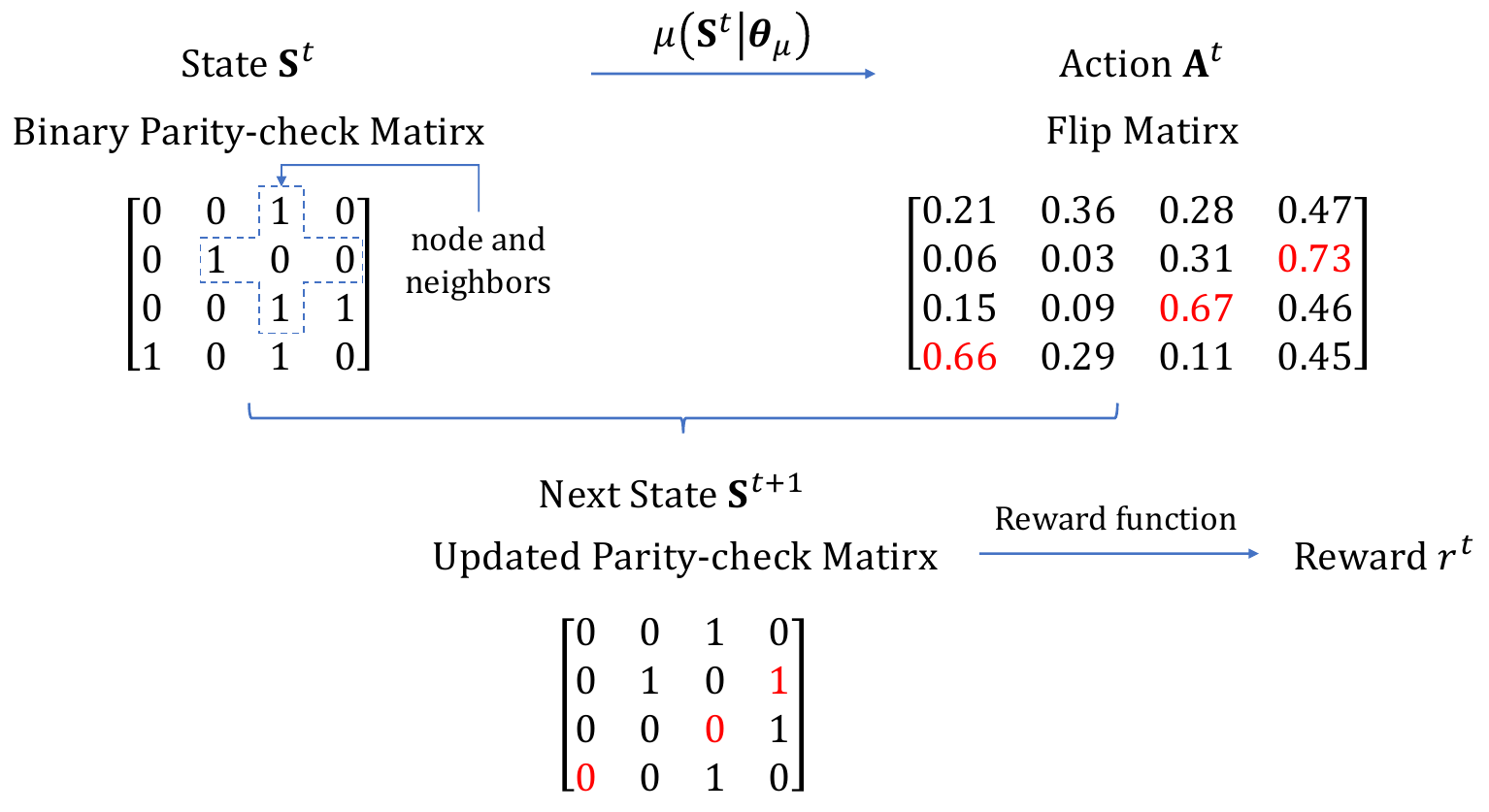}
\vspace{-0.4em}
\caption{An example of the full graph-based MDP generating a parity-check matrix. The flipping threshold $\alpha_f$ is set as $0.5$. }
\vspace{-0.6em}
\label{fig::mdp}
\end{figure}

To address {these} issues, we {consider} a full graph-based MDP {where i}n each step, the agent builds up a complete {lattice graph of the} whole parity-check matrix. 
{Specifically, each matrix element is a node of the graph with links to its upper, lower, right, and left neighbouring elements. This graph is processed by GNN to produce actions, which will be further detailed in Section \ref{sec::enc::GNN}.}
{Figure} \ref{fig::mdp} gives an example of the proposed iterative full graph-based MDP. 
Considering a linear block code $\mathcal{C}(n,k)$, a detailed description of the state representation, the action, and the reward function {of the full graph-based MDP} {is as follows}. 

\subsubsection{States}
We define the state, $\mathbf{S}^t$, {as the parity-check matrix} at time step $t$, which is fully observed by the DRL agent. At the start of each episode, $\mathbf{S}^1$ is initialized by parity-check matrices from {existing} well-designed linear block codes \cite{rptudataset}.
\subsubsection{Actions}
{We define a continuous, high-dimensional action space for link prediction. At each step $t$, the actor network determines the action $\mathbf{A}^t \in \mathbb{R}^{m \times n}$, with all elements scaled to $[0, 1]$.} 
The agent flips elements of the state matrix according to {$\mathbf{A}^t$}. 
The state transition rule is {given by}
\begin{equation} \label{equ::mdp_trans}
    {S}^{t+1}_{j, i} = 
    \begin{cases}
    1 - {S}^t_{j, i}, & \text{if} \ \  {A}^t_{j, i} > \alpha_f, \\
    {S}^t_{j, i}, & \text{otherwise}. \\
    \end{cases}
\end{equation}
{for} ${1} \leq j \leq m$, ${1}  \leq i \leq n$, {where} $\alpha_f$ {is a} predetermined threshold for flipping. 
{Figure} \ref{fig::mdp} shows the generated matrix state before and after an action is taken. 

\subsubsection{Reward} 
\begin{algorithm} [t]
\label{alg::reward}
\small
\caption{Reward Function $R(\mathbf{S}^t, \mathbf{A}^t, \mathbf{S}^{t+1})$}
\SetKwComment{Comment}{$\triangleright$\ }{}
\SetKwInOut{Input}{Input}\SetKwInOut{Output}{Output}
\SetKwFor{For}{for}{do}{endfor} 
\Input{State $\mathbf{S}^t \in \mathbb{F}_2^{m \times n}$, Action $\mathbf{A}^t \in \mathbb{R}^{m \times n}$}
\Output{$r^t \in [0, \infty)$}
Obtain the updated state $\mathbf{S}^{t+1}$ by \eqref{equ::mdp_trans};\\
\uIf{$rank(\mathbf{S}^{t+1}) = m$}{
    Depend on the selected reward method, compute $r^t = r_v + r^t_d$ by \eqref{eq::dec_reward}, or $r^t = r_v + r^t_s$ by \eqref{eq::str_reward};  
}
\Else{$r^t = 0;$}
\end{algorithm}

We use step-wise rewards to guide the behaviour of the DRL agent, which {combine the} validity rewards and {the} coding performance rewards. The reward design is summarised in Algorithm \ref{alg::reward}. {The validity reward is designed to incentives the generation of valid parity-check matrices.} As introduced in Section \ref{sec::lbc}, a valid linear block code must {have a full-rank }parity-check matrix.
{Therefore}, {we assign} {an additional} reward value $r_v$ {to valid matrices with full rank to allow} further performance assessment. 
{In contrast, {the reward is zero if the matrix is invalid.}}

The coding performance reward is designed to evaluate the error-correction capability of the generated code.
We {propose} two types of {coding performance rewards. One is measured from the decoding BER, while another is evaluated directly from the code structure.}

The decoding {BER} reward is {directly} measured from {the simulated BER} with a certain decoder under AWGN channels. 
Given an updated matrix state $\mathbf{S}^{t+1}$ at step $t$, we treat it as a parity-check matrix and obtain {its} corresponding generator matrix $\mathbf{G}^{t+1}$. 
The decoding BER reward value is {obtained} as 
\begin{equation} \label{eq::dec_reward}
    r_d^t = |\ln({\epsilon_b}(\mathbf{S}^{t+1}))|,
\end{equation}
by performing simulations until 10,000 bit errors are collected. 

The code structure reward {estimates} the code performance by examining MHD {and the number of cycles in the Tanner graph}.
As introduced in Section \ref{sec::lbc}, {the larger MHD, the better the code error-correction capability.} 
{We denote t}he MHD of the updated state matrix $\mathbf{S}^{t+1}$ as $d_{\min}(\mathbf{S}^{t+1})$, {which} is obtained by finding {the {codeword from $\mathbf{G}^{t+1}$} with the minimum Hamming weight}.
{On the other hand, to ensure that the designed code is suitable for Tanner graph-based decoding, e.g., BP and EW-GNN,} we incorporate {cycle} analysis into our reward design.
{For the} Tanner graph {of} $\mathbf{S}^{t+1}$, {we represent} the number of $\lambda$-cycles  as $c_\lambda(\mathbf{S}^{t+1})$, {which can be {obtained} using the algorithm in} \cite{counting_cycs}.  
{We consider the impact of both $d_{\min}(\mathbf{S}^{t+1})$ and $c_\lambda(\mathbf{S}^{t+1})$ over the code performance, and propose the code structure reward function as}
\begin{equation} \label{eq::str_reward}
r_s^t = \frac{d_{\min}(\mathbf{S}^{t+1})}{\alpha_{d}} + \frac{\alpha_c}{c_\lambda(\mathbf{S}^{t+1}) + \alpha_c},
\end{equation}
where $\alpha_d$ and $\alpha_c$ are constants used to limit the range of the reward value. 
{The value of $\lambda$ is set to 4 to reduce the number of shortest cycles and decrease the Tanner graph density.}

\vspace{-0.8em}
\subsection{{GNN-based Actor and Critic Networks}} \label{sec::enc::GNN}
\begin{figure}[t]
\centering
\includegraphics[width=0.4\textwidth]{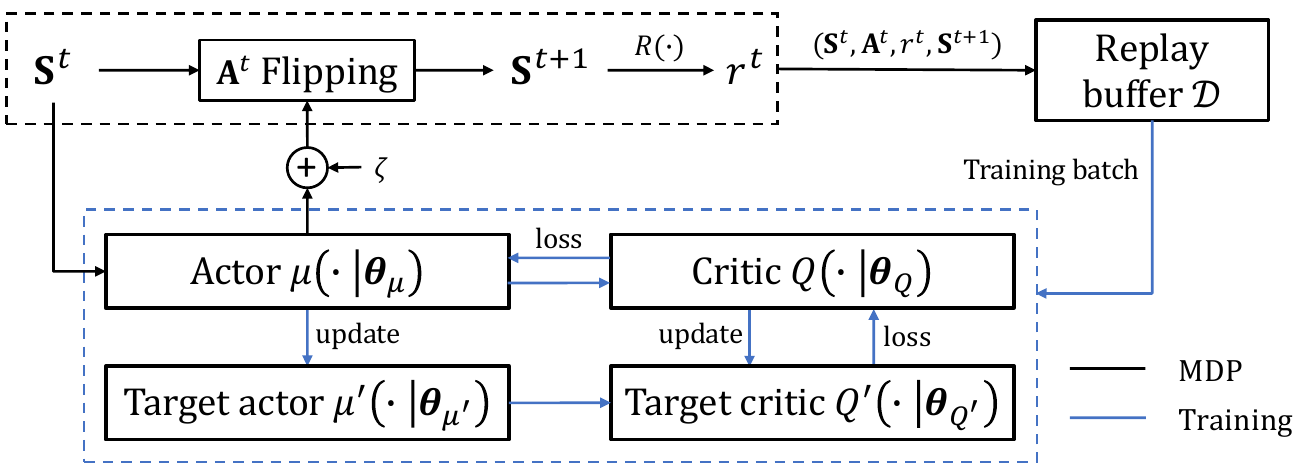}
\vspace{-0.4em}
\caption{The block diagram of the DRL-based code designer.}
\vspace{-1em}
\label{fig::code_designer}
\end{figure}

\vspace{-0.2em}
Based on DDPG, we utilize the actor-critic algorithm for training our model. 
The actor network receives the intermediate matrix state $\mathbf{S}^t$ as input and deterministically map{s} it to an action $\mathbf{A}^t$.
{Then,} {t}he critic network takes the action-state pair as input, and outputs the corresponding Q-value. 
{Figure \ref{fig::code_designer} provides the block diagram of the DRL-based code designer, illustrating the roles of the actor and critic networks in the full graph-based MDP and the training process.}
{In our model, we introduce a lattice graph representation of the state {and} design a message-passing matrix neural network (MPMNN), which is an efficient variant of GNN. 
The proposed MPMNNs are employed in actor and critic networks to extract features from $\mathbf{S}^t$ and the pair of $(\mathbf{S}^t, \mathbf{A}^t)$, respectively.
}

\subsubsection{{Lattice graph representation of {the parity-check} matrix}} 
{By} taking the actor network as an example, we represent the input matrix {$\mathbf{S}^t $} as a {lattice} graph, in which each element of $\mathbf{S}^t$ is treated as a node, {and} its immediate neighbors are the adjacent elements to the left, right, above, and below. 
Notably, $\mathbf{S}^t$ exhibit{s} permutation invariance, {meaning that any permutation of the rows or columns of a parity-check matrix will result in the same code properties due to the linear nature of the codes.} 
Thus, we define the neighbor nodes of boundary elements (i.e. elements in the first row, first column, last row, and last column) cyclically, creating a wrap-around connection. 
An element in the first (or last) row has its upward (or downward) neighbor in the last (or first) row of the same column. Similarly, an element in the first (or last) column has its left (or right) neighbor in the last (or first) column of the same row. 
{The action $\mathbf{A}^t$ can be presented as a lattice graph similar to that for $\mathbf{S}^t$. We note that $\mathbf{A}^t$ and $\mathbf{S}^t$ are of the same dimension, $m\times n$, and they will be processed jointly at the critic network. }

\subsubsection{{Message passing-based matrix neural network}} 

{T}he {input feature of} MPMNN {is denoted as $\mathbf{X} \in \mathbb{R}^{m \times n \times d_0}$}{, which} is a $m \times n$ matrix with each element being a vector of length $d_0$.
In the matrix $\mathbf{X}$, the element at the $j$-th row and the $i$-th column is defined as node $(j, i)$. 
The node feature is directly obtained from the value of corresponding element in the {input} matrix, represented as a $d_0$-dimensional vector $\mathbf{x}_{ji}$. 

Following the message passing process in the MPNN framework \cite{mpnn}, we generate node embeddings iteratively for a total of $L$ layers.
At the $l$-th layer, the hidden embedding of node $(j, i)$, denoted as $\hat{\mathbf{x}}_{ji}^{l}$, is updated based on messages aggregated from its neighbors {according to}
\begin{equation} \label{eq::mpmnn_update}
\begin{split}
\hat{\mathbf{x}}_{ji}^{l+1} &= f_e(\hat{\mathbf{x}}_{ji}^{l}, \ \mathbf{m}_{ji}^{l} \mid \bm{\theta}_e), \\
\mathbf{m}_{ji}^{l} &= \mathrm{agg}(\mathbf{m}_{ji,l}^l, \ \mathbf{m}_{ji,r}^l, \ \mathbf{m}_{ji,u}^l, \ \mathbf{m}_{ji,d}^l),
\end{split}
\end{equation}
where $\hat{\mathbf{x}}_{ji}^{l+1}$ is a $d_e$-dimensional vector.
The parameterized function $f_e(\cdot\mid\bm{\theta}_{e})$ is a multilayer perceptron (MLP) with trainable parameters $\bm{\theta}_{e}$, working as the embedding updating function in a standard MPNN, and
$\mathrm{agg}(\cdot)$ is an aggregation function.
For node $(j,i)$, $\mathbf{m}_{ji,l}^l, \mathbf{m}_{ji,r}^l, \mathbf{m}_{ji,u}^l, \mathbf{m}_{ji,d}^l$ are the messages passed from its neighbors in left, right, up and down directions {at the $l$-th layer}, and {they} are calculated {as}
\begin{equation}\label{eq::mpmnn_message}
\begin{split}
    &\mathbf{m}_{ji,l}^l = f_r(\hat{\mathbf{x}}_{ji}^{l}, \hat{\mathbf{x}}_{ji,l}^{l}\mid\bm{\theta}_{r}), \ \  \ \mathbf{m}_{ji,r}^l = f_r(\hat{\mathbf{x}}_{ji}^{l}, \hat{\mathbf{x}}_{ji,r}^{l}\mid\bm{\theta}_{r}),\\
    &\mathbf{m}_{ji,u}^l = f_c(\hat{\mathbf{x}}_{ji}^{l}, \hat{\mathbf{x}}_{ji,u}^{l}\mid\bm{\theta}_{c}), \ \ \mathbf{m}_{ji,d}^l = f_c(\hat{\mathbf{x}}_{ji}^{l}, \hat{\mathbf{x}}_{ji,d}^{l}\mid\bm{\theta}_{c}),\\
\end{split}
\end{equation}
where {$\hat{\mathbf{x}}_{ji,l}^{l}, \hat{\mathbf{x}}_{ji,r}^{l}, \hat{\mathbf{x}}_{ji,u}^{l}, \hat{\mathbf{x}}_{ji,d}^{l}$ are embeddings of the neighbours of node $(j,i)$ {at the $l$-th layer}.
Note that} $f_r(\cdot\mid\bm{\theta}_{r})$ and $f_c(\cdot\mid\bm{\theta}_{c})$ are two distinct MLPs to process incoming messages from same-row neighbors and same-column neighbors, respectively. 
{These messages passed from neighbors are vectors of length $d_m$.}
At $l=0$, we initialize embeddings {for each node} by applying a linear transformation
\begin{equation} \label{eq::mpmnn_init} 
    \hat{\mathbf{x}}_{ji}^0 = f_{\mathrm{in}}(\mathbf{x}_{ji} \mid \bm{\theta}_{in}), 
\end{equation} 
where $f_{\mathrm{in}} (\cdot \mid \bm{\theta}_{in})$ is a dense layer that maps the $d_0$-dimensional node into a $d_e$-dimensional embedding $\hat{\mathbf{x}}_{ji}^0$, and $d_e > d_0$.

{The structure and flow of the MPMNN are summarised in Algorithm \ref{alg::mpmnn}. 
As can be seen, the {parameters} of the MPMNN {includes} $\bm{\theta}_e$, $\bm{\theta}_r$, $\bm{\theta}_c$, and $\bm{\theta}_{in}$ {in} \eqref{eq::mpmnn_update}-\eqref{eq::mpmnn_init}.}

\begin{algorithm} [t]
\small
\label{alg::mpmnn}
\caption{MPMNN $f_{m}(\mathbf{X} \mid \bm{\theta}_{m})$}
\SetKwComment{Comment}{$\triangleright$\ }{}
\SetKwInOut{Input}{Input}\SetKwInOut{Output}{Output}
\SetKwFor{For}{for}{do}{endfor} 
\Input{Feature matrix $\mathbf{X} \in \mathbb{R}^{m\times n\times d_0}$}
\Output{Embedding matrix $\hat{\mathbf{X}} \in \mathbb{R}^{m\times n\times d_e}$}
{Initialize node embeddings by \eqref{eq::mpmnn_init} for all node $(j,i) \in \mathbf{X}$}
\For{$l \leftarrow 1 : L$}{
\ForEach{node $(j,i) \in \mathbf{X}$}{
    Find embeddings of its neighbors; \\
    Compute messages by \eqref{eq::mpmnn_message};\\
    Update its embedding by \eqref{eq::mpmnn_update};\\
}
}
\end{algorithm}


\subsubsection{{Action Prediction and Critic Evaluation}}
\begin{algorithm} [t]
\small
\label{alg::act_net}
\caption{Actor Network $\mu(\mathbf{S}^t \mid \bm{\theta}_\mu)$}
\SetKwComment{Comment}{$\triangleright$\ }{}
\SetKwInOut{Input}{Input}\SetKwInOut{Output}{Output}
\SetKwFor{For}{for}{do}{endfor} 
\Input{State $\mathbf{S}^t \in \mathbb{F}_2^{m \times n}$}
\Output{Action $\mathbf{A}^t \in \mathbb{R}^{m \times n}$}
Obtain $\mathbf{X}^a = \mathbf{S}^t \in \mathbb{R}^{m\times n\times 1}$; \\
\Comment{Implement MPMNN with input $\mathbf{X}^a$:}
Execute Algorithm \ref{alg::mpmnn};\\
\Comment{Readout phase:}
\ForEach{node $(j,i) \in \mathbf{X}^a$}{
    Compute the action value by \eqref{eq::actor_ro};
}
\end{algorithm}

\begin{algorithm} [t]
\label{alg::q_net}
\small
\caption{Critic Network $Q(\mathbf{S}^t, \mathbf{A}^t\mid \bm{\theta}_Q)$}
\SetKwComment{Comment}{$\triangleright$\ }{}
\SetKwInOut{Input}{Input}\SetKwInOut{Output}{Output}
\SetKwFor{For}{for}{do}{endfor} 
\Input{State $\mathbf{S}^t \in \mathbb{F}_2^{m \times n}$, action $\mathbf{A}^t \in \mathbb{R}^{m \times n}$}
\Output{Q-vlaue $q^t$}
Obtain $\mathbf{X}^c = \mathrm{concat}(\mathbf{S}^t , \mathbf{A}^t ) \in \mathbb{R}^{m\times n\times 2}$;\\
\Comment{Implement MPMNN with input $\mathbf{X}^c$:}
Execute Algorithm \ref{alg::mpmnn};\\
\Comment{Readout phase:}
Compute the Q-value $q^t$ by \eqref{eq::q_ro}. 
\end{algorithm}

We apply {two independent MPMNNs to construct the actor network and the critic network, denoted as $f^a_{m}(\mathbf{X}^a \mid \bm{\theta}^a_{m})$ and $f^c_{m}(\mathbf{X}^c \mid \bm{\theta}^c_{m})$, respectively.}
{In the actor network $\mu(\mathbf{S}^t \mid \bm{\theta}_\mu)$, $\mathbf{S}^t$ is directly used as the input of $f^a_{m}(\cdot \mid \bm{\theta}^a_{m})$, i.e., $\mathbf{X}^a = \mathbf{S}^t \in \mathbb{R}^{m\times n\times 1}$ and $d_0 = 1$.}
After $L$ layers, the readout phase computes the final action value based on the output embeddings, according to
\begin{equation} \label{eq::actor_ro}
    {A}_{j,i}^t = \sigma(f_a(\hat{\mathbf{x}}_{ji}^{L}\mid\bm{\theta}_{a})),
\end{equation}
where $f_{a} (\cdot \mid \bm{\theta}_{a})$ is a dense layer with inputs of dimension $d_e$ and outputs of dimension one. The sigmoid function $\sigma(\cdot)$ is chose{n} to ensure that the action value is constrained within the range $(0, 1)$. 
{The design of the actor network {are} detailed in Algorithm \ref{alg::act_net}, where its parameter $\bm{\theta}_\mu$ is the subsumption of $\bm{\theta}^a_m$ from $f^a_{m}(\mathbf{X}^a \mid \bm{\theta}^a_{m})$ and $\bm{\theta}_a$ from \eqref{eq::actor_ro}.}

In the critic network $Q(\mathbf{S}^t, \mathbf{A}^t\mid \mathbf{\theta}_Q)$, 
{the input of $f^c_{m}(\cdot \mid \bm{\theta}^c_{m})$ is given by $\mathbf{X}^c = \mathrm{concat}(\mathbf{S}^t , \mathbf{A}^t ) \in \mathbb{R}^{m\times n\times 2}$}{, where the $\mathrm{concat}$ operation combines $\mathbf{S}^t$ and $\mathbf{A}^t$ along the last dimension}.
Then we perform the MPMNN over this matrix.
Following $L$ layers, the readout phase calculates the Q-value by aggregating the embeddings of all nodes, as expressed by
\begin{equation} \label{eq::q_ro}
    q^t = {\mathrm{ReLU}}(f_q(\mathrm{agg}(\hat{\mathbf{x}}_{ji}^{L}: 1 \leq j \leq m, 1 \leq i \leq n)\mid\bm{\theta}_{q})),
\end{equation}
where $f_{q} (\cdot \mid \bm{\theta}_{q})$ denotes a dense layer that transforms inputs of size $d_e$ into a single-dimensional output. To obtain non-negative Q-values, we use the rectified linear unit (ReLU) as the activation function. 
The steps {to compute the output of the} critic network {are} summarized in Algorithm \ref{alg::q_net}, {where its parameter $\bm{\theta}_Q$ is the subsumption of $\bm{\theta}^c_m$ from $f^c_{m}(\mathbf{X}^c \mid \bm{\theta}^c_{m})$ and $\bm{\theta}_q$ from \eqref{eq::q_ro}.}

\vspace{-0.6em}
\subsection{Training Algorithm} \label{Sec::enc_train}
\vspace{-0.2em}

The training process of the DRL-based code designer follows the algorithm {detailed} in Algorithm \ref{alg::ddpgenc}. 
During the training phase, we identify the optimal code by selecting the generated parity-check matrix{, represented as $\mathbf{H}^*$,} {which is the state $\mathbf{S}^{t+1}$} that yields the highest reward.

At each time step {$t$}, the agent observes the current state $\mathbf{S}^t$ and generates an action by applying the actor network in Algorithm \ref{alg::act_net}.
To improve the exploration efficiency in our policy, we introduce noise to the actions during the training phase. The {final} action ${\mathbf{A}}^t$ is given by
\begin{equation} \label{eq::action_noised}
    {\mathbf{A}}^t = \mathrm{clip}(\Tilde{\mathbf{A}}^t + \bm{\zeta}, {\alpha_{\mathrm{low}}, \alpha_{\mathrm{high}}}), 
\end{equation}
where $\Tilde{\mathbf{A}}^t  = \mu(\mathbf{S}^t \mid \bm{\theta}_\mu)$ is the output of Algorithm \ref{alg::act_net}.
The exploration noise $\bm{\zeta}$ is sampled from a mean-zero Gaussian distribution. Considering the exploration-exploitation trade-off in RL, we design an epsilon-greedy noise selection method, in which $\bm{\zeta}$ are added to the action with a decaying probability $\epsilon$. 
{Once $\bm{\zeta}$ is added, $\epsilon$ is decayed to $\epsilon = 0.995\epsilon$ with a decaying rate $0.995$, and $\epsilon$ is initialized to be 1}.
The variance of the Gaussian noise, $\sigma_\epsilon^2$, is chosen to be close to the flipping threshold in \eqref{equ::mdp_trans}.
The clip function then limits each element in the noised action $\Tilde{\mathbf{A}}^t + \bm{\zeta}$ to $[\alpha_{\mathrm{low}},\alpha_{\mathrm{high}}]$. 

%

After taking a certain action ${\mathbf{A}}^t$, the agent receives a reward $r_t$ and observes a new state $\mathbf{S}^{t+1}$ according to \eqref{equ::mdp_trans}. The transition tuple $(\mathbf{S}^t, \mathbf{A}^t, r^t, \mathbf{S}^{t+1})$ is stored in a replay buffer $\mathcal{D}$, which is a finite sized cache. 
At each time step, a minibatch of size {$N_B$} {is} uniformly selected from the buffer and used as training samples to optimize the actor and critic networks.
The optimization of the critic network is performed by minimizing the following loss function:
\begin{equation} \label{eq::q_loss}
    L(\bm{\theta}_Q) = \frac{1}{N_B}\sum_{n_B = 1}^{N_B}[y^t - Q(\mathbf{S}^t, \mathbf{A}^t\mid \bm{\theta}_Q)]^2.
\end{equation}
The loss is defined as the difference of the Q-value {and the target value $y^t$}, where the Q-value is obtained by Algorithm \ref{alg::q_net}, and the target value $y^t$ is computed according to Bellman equation \cite{sutton2018rl_intro}:
\begin{equation}
    y^t = r^t + \gamma Q^\prime(\mathbf{S}^{t+1}, \mu^\prime(\mathbf{S}^{t+1} \mid \bm{\theta}_{\mu^\prime}) \mid \bm{\theta}_{Q^\prime}), 
\end{equation}
where $\gamma \in [0, 1]$ is a discounting factor. The target networks, $\mu^{\prime}(\cdot \mid \bm{\theta}_{\mu^\prime})$ and $Q^{\prime}(\cdot \mid \bm{\theta}_{Q^\prime})$, are {replicas} of the actor and critic networks, respectively. 
The parameters of target networks are updated {according to,}
\begin{equation} \label{eq::target_update}
    \begin{split}
        \bm{\theta}_{Q^\prime} &\leftarrow \rho \bm{\theta}_Q + (1-\rho) \bm{\theta}_{Q^\prime},\\
        \bm{\theta}_{\mu^\prime} &\leftarrow \rho \bm{\theta}_\mu + (1-\rho) \bm{\theta}_{\mu^\prime},
    \end{split}
\end{equation}
and
where ${0<}\rho \ll 1$ is a hyper-parameter.
{Meanwhile, }the optimal policy {is trained by maximizing} the {Q-value function,}
\begin{equation} \label{eq::a_loss}
    L(\bm{\theta}_\mu) = - \frac{1}{N_B}\sum_{n_B = 1}^{N_B} Q(\mathbf{S}^t, \mu(\mathbf{S}^t \mid \bm{\theta}_\mu) \mid \bm{\theta}_Q).
\end{equation}
During the training phase, the parameters of the actor and critic networks are updated utilizing the Adam optimizer \cite{adam} with given learning rates $\eta_a$ and  $\eta_c$, respectively.

\begin{algorithm} [t]
\label{alg::ddpgenc}
\small
\caption{The {DRL}-Based {Code Designer}}
\SetKwComment{Comment}{$\triangleright$\ }{}
\SetKwInOut{Input}{Input}\SetKwInOut{Output}{Output}
\SetKwFor{For}{for}{do}{endfor} 
\Input{State initialization matrix $\mathbf{S}^1$, reward function $R(\mathbf{S}^t, \mathbf{A}^t, \mathbf{S}^{t+1})$, number of steps $T_c$}
\Output{parity-check matrix $\mathbf{H}^*$}
Initialize the critic network $Q(\cdot\mid\bm{\theta}_Q)$ and $\mu(\cdot\mid\bm{\theta}_\mu)$ with {random} parameters $\bm{\theta}_Q$ and $\bm{\theta}_\mu$;\\
Initialize target networks by $\bm{\theta}_{Q^\prime} \leftarrow \bm{\theta}_Q, \bm{\theta}_{\mu^\prime} \leftarrow \bm{\theta}_\mu$;\\
Initialize {$r_{max} = 0$ and} an empty replay buffer $\mathcal{D}$;\\
\ForEach{episode}{
    Initialize the environment state $\bm{S}^1$;\\
    \For{$t \leftarrow 1 : T_c $}{ 
        Select action ${\mathbf{A}}^t$ by \eqref{eq::action_noised}, {based on current state $\mathbf{S}^t$};\\
        Observe the new state $\mathbf{S}^{t+1}$ by \eqref{equ::mdp_trans}; \\
        Evaluate reward $r^t$ according to Algorithm \ref{alg::reward}; \\
        \If{$r^t > r_{max}$}{
        $\mathbf{H}^* = \mathbf{S}^{t+1}$;\\
        $r_{max} = r^t$;\\
        }
        Store the transition $(\mathbf{S}^t, {\mathbf{A}}^t, r^t, \mathbf{S}^{t+1})$ in $\mathcal{D}$;\\
        \If{size($\mathcal{D}) \geq N_B$}{
        Randomly sample a batch of $N_B$ transitions from $\mathcal{D}$;\\
        Update $\bm{\theta}_Q$ by minimizing the loss in \eqref{eq::q_loss};\\
        Update $\bm{\theta}_\mu$ by minimizing the loss in \eqref{eq::a_loss};\\
        Update $\bm{\theta}_{Q^\prime}$ and $\bm{\theta}_{\mu^\prime}$ by \eqref{eq::target_update};\\
        }
    }
}
\end{algorithm}

\vspace{-0.6em}
\section{EW-GNN Decoder} \label{sec::EW-GNN}
\vspace{-0.2em}
{GNN is capable of learning well a wide range of practical models based on graphs, which is justified by the \textit{algorithmic alignment} framework \cite{algorithm_aligen}.
GNN can align to the BP decoding algorithm,  because 1) BP {decoding is a} node classification problem on the Tanner graph and 2) both GNN and Tanner graph are permutation-invariant due to the linear property of codes \cite[Section 2]{algorithm_aligen}. 
In this section, we propose an EW-GNN decoder operating on the Tanner graph, which is different from the lattice graph used by DRL code designer in Section \ref{sec::DRL_code_designer}}.

\vspace{-0.6em}
\subsection{GNN Decoder with Edge Weights}  \label{sec::EW-GNN_structure}
\begin{figure}[t]
\centering
\includegraphics[width=0.45\textwidth]{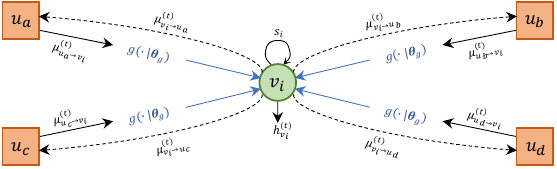}
\vspace{-0.2em}
\caption{Illustration of updating the node embedding in the GNN-based decoder. The variable node $v_i$ receives messages from $\mathcal{M}(v_i)=\{u_a, u_b, u_c, u_d\}$. {$g(\cdot\mid\bm{\theta}_g)$ is a trainable neural network.}}
\vspace{-1em}
\label{fig::emb}
\end{figure}

\vspace{-0.2em}
Considering a bipartite {Tanner} graph $ \mathcal{G}=(\mathcal{V}, \mathcal{U}, \mathcal{E})$ with directed edge messages, as introduced in Section \ref{Sec::Preliminaries::TG}, 
we define the node feature of variable node $v_i$ as the received LLR information\footnote{GNN can apply different update and message functions as per applications.
In the GNN decoder, the node classification only performs on variable nodes representing codeword bits. Thus, we only assign features and embeddings to variable nodes for simplicity.}, $\ell_i$ given by \eqref{equ::llr}.
In each iteration, the hidden node embedding of $v_i${, denoted as $h_{v_i}^{(t)}$, is updated based on incoming edge messages $\mu_{u \to v}^{(t)}$, which is iteratively generated from $\mu_{v \to u}^{(t-1)}$.}
Figure \ref{fig::emb} gives an example of the updating process. 
Since $n$ variable nodes correspond to $n$ codeword bits, the GNN decoder finally obtains ${\hat{c}_{i}}$ from $h_{v_i}^{(T)}$, for $v_i \in \mathcal{V}$, {after} $T$ {iterations}.

In the proposed GNN decoder, we introduce ``weight'' to the edge messages to improve the decoding performance.
{Intuitively, by reducing the weights of} unreliable edge messages
generated from small cycles in the Tanner graph, {it is possible to} improve the  accuracy of the output codeword estimate $\hat{\mathbf{c}}$. 

The weight is a multiplicative correction factor \cite{yazdani2004cycleBP}  working on the edge message from parity-check nodes to variable nodes.
Let us denote the weight of $\mu_{u_j \to v_i}^{(t)}$ in the $t$-th iteration as $w_{u_j \to v_i}^{(t)} \in \mathbb{R}$.
By employing an MLP, $g(\cdot\mid\bm{\theta}_g)$ with trainable parameters $\bm{\theta}_g$, the weight $w_{u_j \to v_i}^{(t)}$ is learned from input features that can reflect the reliability of node and edge information, which is defined as
\begin{equation} \label{equ::ewgnn_weight}
\small
\begin{split}
    w_{u_j \!\to v_i}^{(t)} \!\!=\! g(|\mu_{u_j \!\to v_i}^{(t)}\!|,  \delta^{\!(t)}(\mu_{u_j \!\to v_i}\!),
    \delta^{\!(t\!-\!1)}(\mu_{v_i \!\to u_j}\!), 
    \delta^{\!(t\!-\!1)}({h}_{v_i}\!)
    \mid\bm{\theta}_g
    ).
\end{split}
\end{equation}
In \eqref{equ::ewgnn_weight}, $|\mu_{u_j \to v_i}^{(t)}|$ represents the reliability of edge message $\mu_{u_j \to v_i}^{(t)}$, because $\mu_{u_j \to v_i}^{(t)}$ can be regarded as an extrinsic LLR (see \cite{LLRbp}). 
Then, {$\delta^{(t)}(x)$} is a residual function given by
\begin{equation} \label{equ:ressidual}
    \delta^{(t)}(x) = |x^{(t)}-x^{(t-1)}|.
\end{equation}
Thus, $\delta^{(t)}(\mu_{u_j \to v_i})$, $ \delta^{(t-1)}(\mu_{v_i \to u_j})$, and $\delta^{(t-1)}({h}_{v_i})$ represent the residual value of $\mu_{u_j \to v_i}$, $\mu_{v_i \to u_j}$, and ${h}_{v_i}$, respectively.
They are also measurements of the reliabilities \cite{info_dec_ldpc}; precisely, a lower residual value is, 
a more reliable node/edge will be.


Next, we explain the algorithm of EW-GNN with the help of weight $w_{u_j \to v_i}^{(t)}$.
At the $t$-th iteration, EW-GNN first updates the edge message from $u_j$ to $v_i$ according to
{\small
\begin{equation} \label{equ::ewgnn_utov}
     \mu_{u_j \to v_i}^{(t)} = \ln 
     \frac{{\mathrm{clip}}\left(
     1+ 
     \prod_{v \in \mathcal{M}(u_j)\setminus v_i} 
     \tanh \left(
     \frac{\mu_{v \to u_j}^{(t-1)}}{2}
     \right),
     \alpha, 2-\alpha \right)}
     {\mathrm{clip}\left(
     1- 
     \prod_{v \in \mathcal{M}(u_j)\setminus v_i} 
     \tanh \left(
     \frac{\mu_{v \to u_j}^{(t-1)}}{2} 
     \right),
     \alpha, 2-\alpha \right)}
\end{equation}
}where $\alpha$ is a model parameter less than $10^{-7}$.
We note that the message update function \eqref{equ::u_to_v} in BP is {not applicable} to neural-network-based decoders, because $\tanh^{-1}(\cdot)$ {is unbounded} and introduces non-differentiable singularities, leading to a non-convergence training process. Therefore, {we modify} $\tanh^{-1}(\cdot)$ with the clip function 
to avoid the aforementioned issue. 
Compared to the Taylor series expansion of $\tanh^{-1}(\cdot)$ used in \cite{hyperbp}, the clip function in \eqref{equ::ewgnn_utov} introduces virtually no extra complexity. 

After $\mu_{u_j \to v_i}^{(t)}$ is obtained for $u_j \in \mathcal{U}$ and $v_i \in \mathcal{M}(u_j)$, the updated edge message from $v_i$ to $u_j$ is given by\footnote{$M_t\!(\mu_{u_j \!\to v_i}^{(t)}\!, h_{v_i}^{(t\!-\!1)}) \!=\! w_{u_j \!\to v_i}^{(t)} \mu_{u_j \!\to v_i}^{(t)}$ as $w_{u_j \!\to v_i}^{(t)}$ is a function of $h_{v_i}^{(t\!-\!1)}\!\!.$}
\begin{equation}\label{equ::ewgnn_vtou}
    \mu_{v_i \to u_j}^{(t)} = \ell_i + \sum_{u \in \mathcal{M}(v_i)\setminus u_j}w_{{u \to v_i}}^{(t)} \cdot \mu_{u \to v_i}^{(t)}.
\end{equation} 

Then, {the node embedding $h_v^{(t)}$ is updated by aggregating weighted edge messages from all neighbor nodes} {according to}
\begin{equation} \label{equ::ewgnn_emb}
     {h}_{v_i}^{(t)} = \ell_i + \sum_{u \in \mathcal{M}(v_i)}w_{{u \to v_i}}^{(t)} \cdot \mu_{u \to v_i}^{(t)}. 
\end{equation}

At $t=0$, the node embeddings $h_{v_i}^{(0)}$ and edge messages $\mu_{v_i \to u}^{(0)}$ are all initialized to corresponding input features of node $v_i$, i.e., $\ell_i$ given in \eqref{equ::llr}. Also, all residuals are initialized to 0.
{Meanwhile}, at the $t$-th iteration, EW-GNN uses node embeddings and edge messages propagated from previous two iterations to update $h_v^{(t)}$.

After $T$ iterations, we obtain the estimated codeword $\hat{\mathbf{c}}$ from the node embedding ${h}_{v_i}$ according to the following rule:
\begin{equation} \label{equ::ewgnn_readout}
    \hat{c}_i = 
    \begin{cases}
    1, & {h}^{(T)}_{v_i} \leq 0,\\
    0, & {h}^{(T)}_{v_i} > 0.\\
    \end{cases}
\end{equation}

\begin{algorithm} [t]
\small 
\caption{Training and Inference of EW-GNN}
\label{alg::ewgnn}
\SetKwComment{Comment}{$\triangleright$\ }{}
\SetKwInOut{Input}{Input}\SetKwInOut{Output}{Output}
\SetKwFor{For}{for}{do}{endfor} 
\Comment{Training Phase:}
\Input{$\mathbf{G}$, $\mathbf{H}$, training batch size $N_B$, training SNR $[\,\gamma_{\min}, \gamma_{\max}]\,$, number of iterations $T$, learning rate $\eta$ \\}
\Output{well-trained $g^*(\cdot\mid\bm{\theta}^*_g)$}
Create the network topology, $ \mathcal{G}=(\mathcal{V}, \mathcal{U}, \mathcal{E})$ from $\mathbf{H}$;\\
\ForEach{epoch}{
    \For{$n_B \leftarrow 1 : N_B $}{
        {Encode a random message $\mathbf{b}$ to $\mathbf{c} = \mathbf{b} \mathbf{G} $};\\
        Obtain the noisy signal $\mathbf{y}$ with the AWGN channel at a random SNR $\gamma \in [\,\gamma_{\min},\gamma_{\max}]\,$;\\
        Obtain the received LLR vector $\bm{\ell}$ by \eqref{equ::llr};\\
        Initialize $h_{v_i}^{(0)}\!=\! \ell_i$, $\mu_{v_i \!\to u}^{(0)} \!=\! \ell_i$, $\mu_{u \!\to v} ^{(0)} \!=\! 0$, $\delta^{(0)}\!(\cdot) \!=\! 0$;\\
        \For{$t \leftarrow 1 : T$}{
            \ForEach{$e_{i,j} \in \mathcal{E}$}{
            Update edge message $\mu_{u_j \to v_i}^{(t)}$ by \eqref{equ::ewgnn_utov};\\
            Update weight $w_{u_j \to v_i}^{(t)}$ by \eqref{equ::ewgnn_weight};\\
            Update edge message $\mu_{v_i \to u_j}^{(t)}$ by \eqref{equ::ewgnn_vtou};\\
            }
            \ForEach{$v_i \in \mathcal{V}$}{
            Update node embedding $h_{v_i}^{(t)}$ by \eqref{equ::ewgnn_emb};
            }
        }
        Calculate the multiloss value $L_{n_B}$ by \eqref{equ::lossfuncion};
    }
    Update $\bm{\theta}_g$ with the loss  $L(\bm{\theta}_g)=\frac{1}{N_B}\sum_{n_B=1}^{N_B}L_{n_B}$;
}
\BlankLine
\Comment{Inference Phase:}
\Input{LLRs of the received codewords $\bm{\ell}\in \mathbb{R}^n$, parity-check matrix $\mathbf{H^\prime}$, number of iterations $T^\prime$}
\Output{estimated codewords $\hat{\mathbf{c}}$}
Create the network topology, $ \mathcal{G}^\prime=(\mathcal{V}^\prime, \mathcal{U}^\prime, \mathcal{E}^\prime)$ from $\mathbf{H}^\prime$;\\
Execute line{s 7-17} on $ \mathcal{G}^\prime$  with  $g^*(\cdot\mid\bm{\theta}^*_g)$ for $T'$ iterations;\\
Obtain the estimated codewords $\hat{\mathbf{c}}$ according to \eqref{equ::ewgnn_readout}.
\vspace{-0.1em}
\end{algorithm}

\vspace{-1.2em}
\subsection{Training {Algorithm}}
\vspace{-0.2em}
{The algorithm of the proposed EW-GNN decoder is summarized in Algorithm \ref{alg::ewgnn}, which includes the training phase and {the} inference phase.} 
The goal of the training phase is to find an optimal $g(\cdot\mid\bm{\theta}_g)$  that minimizes the difference between the transmitted codeword and its estimate. 
For channel coding, a large amount of data samples can be easily obtained to perform the end-to-end training  \cite{deepcoding}.
The label data, i.e., the transmitted codewords, are generated by encoding a random binary message. 
By imposing an AWGN with power $\sigma_n^2$ over the symbols of codewords, received LLRs can be computed according to \eqref{equ::llr}, which are used as the input data of EW-GNN. To improve the robustness of EW-GNN, we perform the training process with diverse SNRs uniformly distributed in the interval $[\gamma_{\min}, \gamma_{\max}]\,$. 

Since the channel decoding can be considered as a bit-wise binary classification task, we apply the average binary cross-entropy (BCE) per bit per iteration as the loss function, i.e., 
\begin{equation} \label{equ::lossfuncion}
    L = -\frac{1}{n T}\sum_{t=1}^{T}\sum_{i=1}^{n}{c_i \log(p^{(t)}_{v_i})+(1-c_i) \log(1-p^{(t)}_{v_i})},
\end{equation}
which is the multiloss variant \cite{nbp2018} of a regular BCE loss function. 
In \eqref{equ::lossfuncion}, $p^{(t)}_{v_i}$ denotes the probability $\mathrm{Pr}(c_i = 1)$ estimated from ${h}_{v_i}^{(t)}$ of $v_i$ at iteration $t$, which is given by {$p^{(t)}_{v_i}=1/(1+\exp({{h}_{v_i}^{(t)}})) \in (0,1)$.}
During the training phase, $g(\cdot\mid\bm{\theta}_g)$ parameters are updated utilizing the Adam optimizer with a given learning rate $\eta$ \cite{adam}, {and $L_{n_B}$ in Algorithm \ref{alg::ewgnn} is the loss of the $n_B$-th sample in the batch of size $N_B$.}

\vspace{-0.6em}
\section{Auto-Encoder with {Iterative Training} }
\vspace{-0.2em}

\label{sec::auto_encoder}
\begin{figure}[t]
\centering
\includegraphics[width=0.43\textwidth]{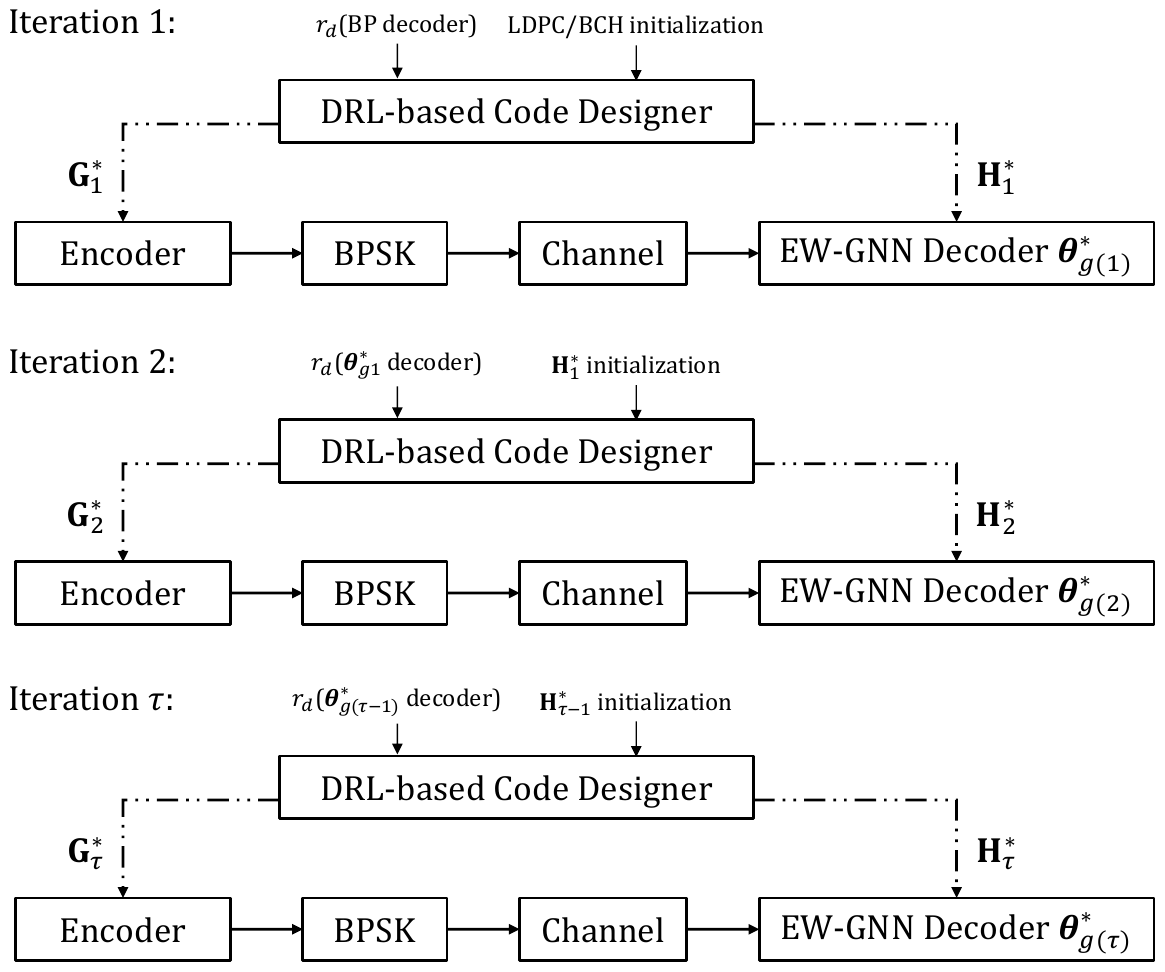}
\vspace{-0.4em}
\caption{{The iterative training phase of the proposed GNN-DRL auto-encoder architecture with $\tau$ training iterations.}}
\vspace{-1em}
\label{fig::ae}
\end{figure}


An auto-encoder framework {for linear block code} provides a communication scheme in which both the encoder and the decoder are implemented by machine learning methods, and {they} can be optimized {jointly}.
{In our auto-encoder model, the DRL-based code designer generates an optimal code with an initial decoder, e.g., BP. The EW-GNN decoder is then trained based on this code. Subsequently, the DRL-based code designer generates a new optimal code with the trained EW-GNN decoder, and the EW-GNN decoder is fine-tuned based on the new code. The encoder and decoder will be trained in this manner for several iterations.}
{We employ this iterative training method because} the direct joint optimization of a DRL-based model and a supervised learning-based model is {difficult} due to the 
vast combined parameter space of two models.
{On the other hand,} the reward-based, unsupervised-learned DRL code designer and the loss-based, supervised-learned EW-GNN decoder have completely different training process{es}.

The {detailed} iterative training process of the proposed auto-encoder framework is illustrated in Fig. \ref{fig::ae}. 
At the first iteration, a DRL-based code designer is independently trained using the BER reward measured by BP decoding and outputs the learned code $\mathbf{H}^*_{1}$. 
{C}onventional linear block codes, such as LDPC and BCH, are chosen as the state initialization matrices {at the first iteration}.
The EW-GNN decoder is then trained based on $\mathbf{H}^*_{1}$ {and its corresponding generator matrix $\mathbf{G}^*_{1}$} according to Algorithm \ref{alg::ewgnn}, providing a well-trained decoder network, which can be represented by its parameters $\bm{\theta}_{g(1)}$.
{In the second training iteration}, the $\bm{\theta}_{g(1)}$ decoder is employed to measure the decoding BER reward, and the state {of the DRL-based code designer is initialized as $\mathbf{H}^*_{1}$}.
{Then, the code designer generates} $\mathbf{H}^*_{2}$, which are used to train a new EW-GNN decoder $\bm{\theta}_{g(2)}$.
This {training }process is iteratively repeated for a few {iterations}.
After $\tau$ iterations, {our auto-encoder framework produces} an expertly designed linear block code represented by $\mathbf{H}^*_\tau$, accompanied by an optimally trained EW-GNN decoder with parameters $\mathbf{\theta}^*_{g(\tau)}$. These components can be jointly employed to assemble a channel coding system with superior error-control capabilities.
{Our auto-encoder has offline generation of codes and associated decoders, as the on-the-fly design of {codes} is impractical due to the extensive complexity DRL model training.}
Once the auto-encoder generates an optimal linear block code from the DRL-based code designer, {$\mathbf{H}^*_\tau$ and} $\mathbf{G}^*_\tau$ can be stored to enable fast online encoding by \eqref{equ::enc} {and} fast online decoding by executing the inference phase in Algorithm \ref{alg::ewgnn}.
The EW-GNN algorithm has a low inference complexity that increases linearly with the code length\cite{ewgnn}{; thus, its decoding complexity is similar to BP.}

Our auto-encoder framework guides the code matrix design based on a {trainable} decoding method, allowing the encoder and decoder to reciprocally train toward optimal performance;
{Moreover, the state initialization and the reward function are continuously updated, improving the exploration efficiency.}

\vspace{-0.8em}
\section{Experiments and Results} \label{Sec::experiments}
\vspace{-0.4em}

{This section presents} the results obtained {from} our proposed algorithms.
We first evaluate the performance of the DRL-based code designer and the EW-GNN decoder separately. 
For the code designer, we examine the BER performance of the {learned} codes, by using them to encode {messages} over AWGN channel with BP and MLD {decoders}. 
Meanwhile, the {BER performance} of the proposed {EW-GNN} decoder is {evaluated} by using it to decode classic {{analytically} designed} codes.
{Subsequently,} we {will} demonstrate the performance of the proposed auto-encoder framework. All BER results are measured at each SNR {by performing simulations until 10,000 bit errors are collected.} 
{For different coding schemes, we define} \textit{coding gain} as the difference in SNRs required to achieve the same target BER {at high SNRs}.

\vspace{-0.6em}
\subsection{DRL-based Neural Code Designer}\label{sec::enc_simulation}
\vspace{-0.2em}
{We evaluate the BER of the codes generated by our DRL-based code designer, }as outlined in Algorithm \ref{alg::ddpgenc}.
{W}e employ two different types of reward functions {to} generate two distinct codes $\mathbf{H}^*_d$ and $\mathbf{H}^*_s$.
Specifically, $\mathbf{H}^*_d$ {is} produced using the decoding BER reward {in \eqref{eq::dec_reward}} {with a 8-iteration BP decoder over $6~\mathrm{dB}$ AWGN}, and $\mathbf{H}^*_s$ {is} produced using the code structure reward in \eqref{eq::str_reward}.
{Their} BER {performance} is evaluated with the BP decoder {with 8 iterations} and MLD. 
{BP is an efficient suboptimal decoder designed for LDPC codes, while MLD can fully exploit the error-correction capability of any codes.}
This {approach} allows us to test the designed codes under both optimal and suboptimal decoders. 
{The training {algorithm} of the DRL-based code design{er} is implemented using TensorFlow, with hyper-parameters detailed in Table \ref{tab:enc_parameters}.}

\begin{table}[t]
\caption{Hyper-parameters of the proposed {DRL-based} code designer}
\label{tab:enc_parameters}
\renewcommand\arraystretch{1.2}   
\centering
\begin{tabular}{lll}
\hline
\multicolumn{2}{l}{Parameters} & Value \\ \hline \hline
\multirow{3}{*}{Environment} & State initialization $\mathbf{S}^1$ & CCSDS $(32,16)$ LDPC \\ 
 & Flipping threshold $\alpha_f$ & $0.3$ \\
 & {$\mathbf{A}^t$ range $[\alpha_{\mathrm{low}},\alpha_{\mathrm{high}}]$} & {$[0, 1]$} \\
 & Step number $T_c$ & $25$ \\ \hline
\multirow{2}{*}{BER  reward} & Decoding method & BP with 8 iterations \\
 & Decoding SNR & $6 ~\mathrm{dB}$ \\ \hline
\multirow{2}{*}{Structure reward} & $\alpha_d$ & $8$ \\
 & $\alpha_c$ & $500$ \\ \hline
\multirow{5}{*}{MPMNN} & Embedding dimension {$d_e$} & 10 \\
 & Message dimension & $10$ \\
 & MPMNN layers & $3$ \\
 & MLP layers & $3$ \\
 & MLP hidden units & $40$ \\ \hline
\multirow{1}{*}{Noise} & Gaussian   noise factor $\sigma_\epsilon$ & $0.3$\\ \hline
\multirow{5}{*}{Training} & Learning rate{s} $\eta_a$ {and $\eta_c$} & $0.001$ \\
 & Discount factor $\gamma$ & $0.99$ \\
 & Target update factor $\rho$ & $0.001$ \\
 & Batch size $N_B$ & $128$ \\ \hline
\end{tabular}
\end{table}

\begin{figure}[t]
\centering
\includegraphics[width=0.39\textwidth]{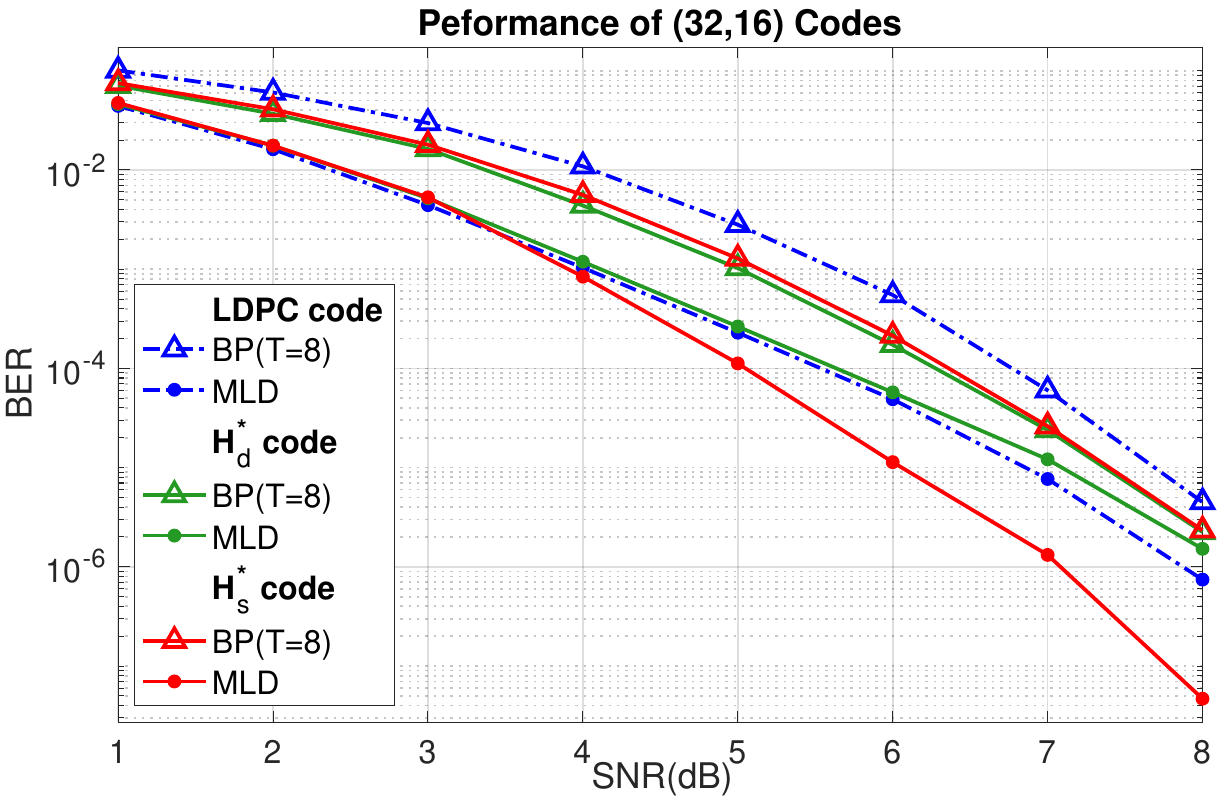}
\vspace{-0.4em}
\caption{BER performance for designed $(32, 16)$ linear block codes.}
\vspace{-1.2em}
\label{fig::enc_3216}
\end{figure}

\begin{table}[t]
\caption{Code properties of designed $(32, 16)$ linear block codes}
\label{tab:property_result}
\renewcommand\arraystretch{1.2}   
\centering
\begin{tabular}{llll}
\hline
Code properties    & LDPC    & $\mathbf{H}^*_d$   & $\mathbf{H}^*_s$   \\ \hline \hline
MHD                & $4$       & $3$      & $5$      \\ \hline
Number of 4-cycles & $178$     & $52$     & $72$     \\ \hline
Number of 6-cycles & $2090$    & $487$   & $685$   \\ \hline
Density            & $0.250$ & $0.186$ & $0.193$ \\ \hline
BER with BP {at} 5dB  & $2.81e-3$ &$1.03e-3$ & $1.28e-3$ \\ \hline
BER with ML {at} 5dB  & $2.29e-4$ &$2.64e-4$ & $1.12e-4$\\ \hline
\end{tabular}
\vspace{-1em}
\end{table}

As illustrated in Fig. \ref{fig::enc_3216}, the {DRL-based code designer generates $\mathbf{H}^*_d$ and $\mathbf{H}^*_s$ with parameters $k=16, n=32$} and {they are} compared with the traditional CCSDS $(32,16)$ LDPC \cite{ccsds2015short}.
{As shown,} both $\mathbf{H}^*_d$ and $\mathbf{H}^*_s$, can achieve better {BER} performance than the LDPC {code}.
When {decoded by BP}, $\mathbf{H}^*_d$ achieves the best decoding performance among all {the} counterpart{s}, and {has $0.65 ~\mathrm{dB}$ more coding gain than} the {LDPC code} at high SNRs around $6 ~\mathrm{dB}$.
Although the decoding BER reward value is measured {only at the SNR of} $6 ~\mathrm{dB}$, 
the learned $\mathbf{H}^*_d$ demonstrates superior performance over a wide range of SNRs.
Under {MLD}, $\mathbf{H}^*_s$ {achieves} better BER performance {than} LDPC, {with} an increased coding gain {of} $ 0.83 ~\mathrm{dB}$ at higher SNRs. This validates the {performance} of the proposed DRL-based code designer {in learning} {good} codes {with superior error-correction capability.}

{We further} examine the algebraic structure properties of these $(32,16)$ codes, as presented in Table \ref{tab:property_result}.
{As shown in the table,} the MHD of the DRL-generated matrices $\mathbf{H}^*_d$ and $\mathbf{H}^*_s$ are 3 and 5, respectively, while the CCSDS $(32,16)$ LDPC code has an MHD of 4.
{As a result,} $\mathbf{H}^*_d$ {exhibits inferior performance} compared to {the} LDPC {code} when {decoded using the} {MLD}, {while $\mathbf{H}^*_s$ allows an excellent MLD performance.}
Table \ref{tab:property_result} also {demonstrates} that the density of {these} parity-check {matrices} and the number of short cycles in {their} associated Tanner graphs.
Compared to the benchmark LDPC, the learned $\mathbf{H}^*_d$ and $\mathbf{H}^*_s$ are both more sparse and have less number of {cycles of length 4}.
{Consequently, {both} $\mathbf{H}^*_d$ and $\mathbf{H}^*_s$ outperform the LDPC code under the BP decoding.}

{We note that t}he BP-decoding BER reward function tends to produce matrices with fewer short cycles and lower density, but this {leads} to poor MHD performance {of $\mathbf{H}^*_d$}. 
{This can be addressed by the auto-encoder framework proposed in Section \ref{sec::auto_encoder},} {whose performance will be} demonstrated in Section \ref{sec::ae_performance}.

\vspace{-1em}
\subsection{EW-GNN Decoder}\label{sec::dec_simulation}
\vspace{-0.2em}
In this subsection, we evaluate the error-correction performance and the scalability of the proposed EW-GNN for short BCH codes and LDPC codes. 
The BP and NBP decoders \cite{nbp2018} are considered as benchmarks for performance comparison. 
The decoding performance of these algorithms {is} measured by BER at various SNRs. 
The training of the EW-GNN algorithm is implemented using TensorFlow with hyper-parameters in Table \ref{tab::parameters}. 
\begin{table}[tb]
\renewcommand\arraystretch{1.2}   
\centering
\caption{Hyper-parameters of the proposed EW-GNN decoder}
\label{tab::parameters}
\begin{tabular}{lcc}
\hline
\multirow{1}{*}{Parameters} 
       &  \multicolumn{1}{c}{BCH codes} & LDPC codes \\ \hline \hline
    Clip factor $\alpha$  & \multicolumn{1}{c}{$10^{-32}$} &  $10^{-7}$ \\ \hline
    SNR range  $[\gamma_{\min}, \gamma_{\max}]$    & \multicolumn{1}{c}{$[3 ~\mathrm{dB}, 8 ~\mathrm{dB}]$} & $[1 ~\mathrm{dB}, 8 ~\mathrm{dB}]$ \\ \hline
    Batch size  $N_B$            & \multicolumn{1}{c}{2000} &  4000 \\ \hline
    Learning rate {$\eta$}   & \multicolumn{2}{c}{$10^{-3}\sim 10^{-5}$}    \\ \hline
     Number of iterations $T$            & \multicolumn{2}{c}{8}    \\ \hline
     MLP layers             & \multicolumn{2}{c}{3}    \\ \hline
     MLP hidden units             & \multicolumn{2}{c}{32}    \\ \hline
     ELU factor $\beta$          & \multicolumn{2}{c}{1.0}    \\ \hline
\end{tabular}
\vspace{-1em}
\end{table}

\begin{figure*}[bt]
    \centering
    \subfloat[$k=51$]{\includegraphics[width=0.315\textwidth]{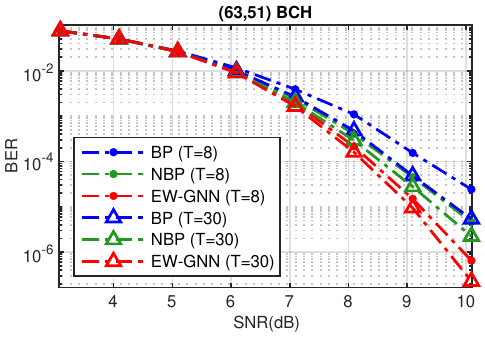}\label{fig::6351bch}}
    \hfill
    \subfloat[$k=45$]{\includegraphics[width=0.315\textwidth]{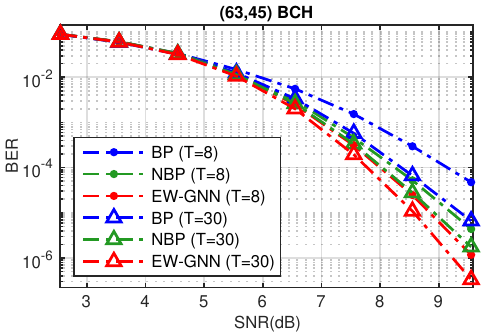}\label{fig::6345bch}}
    \hfill
    \subfloat[$k=36$]{\includegraphics[width=0.315\textwidth]{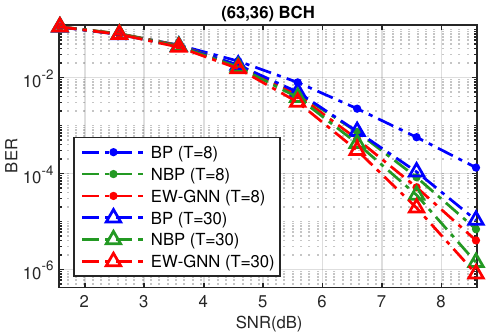}\label{fig::6336bch}}
    \vspace{-0.4em}
    \caption{BER performance for BCH codes of length $n=63$. EW-GNN is only trained with the $(63,51)$ BCH code and $T=8$.}
    \vspace{-1em}
    \label{fig::bch_results}
\end{figure*}

\begin{figure*}[t]
    \centering
    \subfloat[$n=32$]{\includegraphics[width=0.315\textwidth]{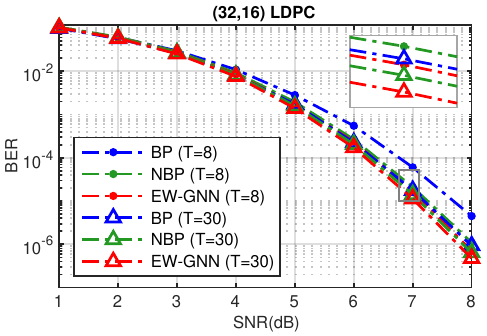}\label{fig::3216ldpc}}
    \hfill
    \subfloat[$n=128$]{\includegraphics[width=0.315\textwidth]{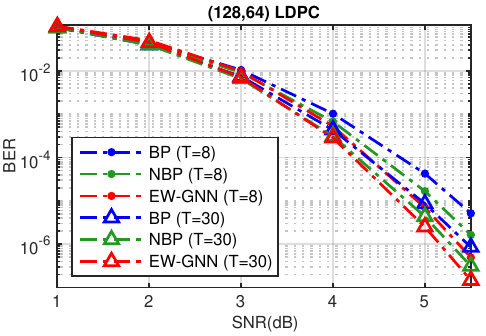}\label{fig::12864ldpc}}
    \hfill
    \subfloat[$n=256$]{\includegraphics[width=0.315\textwidth]{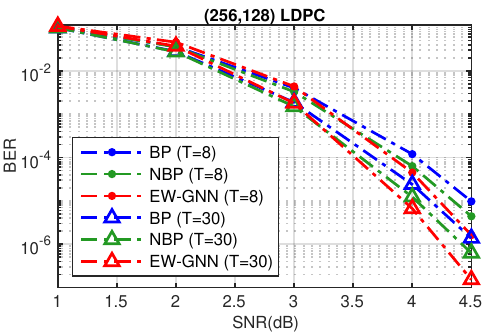}\label{fig::256128ldpc}}
    \vspace{-0.4em}
    \caption{BER performance for LDPC codes at rate $1/2$. EW-GNN is only trained with the $(32,16)$ LDPC code and $T=8$.}
    \vspace{-1em}
    \label{fig::ldpc_results}
\end{figure*}

\subsubsection{BCH Codes}
We first train an EW-GNN decoder for the $(63,51)$ BCH code with $T=8$ decoding iterations, and consider two inference scenarios with different number of decoding iterations, $T=8$ and $T=30$. 

As shown in Fig. \ref{fig::6351bch}, EW-GNN outperforms BP and NBP in terms of BER for $(63,51)$ BCH code with both $T=8$ and $T=30$.  When $T=8$, our proposed EW-GNN decoder has $1.2 ~\mathrm{dB}$ and $0.62 ~\mathrm{dB}$ larger coding gains compared to the BP and NBP algorithms, respectively, at high SNRs.
Although EW-GNN is trained {in the scenario} with $T=8$, its BER performance remains superior with a larger number of decoding iterations. 
As shown, when $T=30$, the EW-GNN decoder outperforms the NBP decoder by $0.61 ~\mathrm{dB}$.
It is worth noting that our proposed decoder with $T=8$ has a better BER performance than the NBP decoder with $T=30$.

We directly apply the EW-GNN model trained for the $(63,51)$ BCH code with $T=8$ to decode $(63,45)$ and $(63,36)$ BCH codes. 
Note that the Tanner graph of $(63,51)$ BCH codes has much fewer edges (i.e., smaller network size) than those of $(63,45)$ and $(63,36)$ BCH codes.
As can been seen in Fig{s}. \ref{fig::6345bch} and \ref{fig::6336bch}, the EW-GNN algorithm still achieves better BER performance than the BP and NBP algorithms for $(63,45)$ and $(63,36)$ BCH codes.
For example, when decoding the $(63,36)$ BCH code with $T=30$, the proposed EW-GNN increases the coding gain by $0.8 ~\mathrm{dB}$ and $0.2 ~\mathrm{dB}$ compared to BP and NBP respectively.

\subsubsection{LDPC Codes}
We train the EW-GNN decoder for the half-rate CCSDS $(32,16)$ LDPC code \cite{ccsds2015short}  with the number of decoding iterations $T=8$.
As depicted in Fig. \ref{fig::3216ldpc}, the proposed EW-GNN algorithm  achieves the best decoding performance among all its counterparts in decoding the $(32,16)$ LDPC code.
Specifically, when $T=30$, EW-GNN improves the coding gains of BP and NBP by $0.21 ~\mathrm{dB}$ and $0.1 ~\mathrm{dB}$ at high SNRs, respectively.

To investigate the scalability of EW-GNN to different code lengths, the EW-GNN model trained for the $(32,16)$ LDPC codes is directly applied to longer LDPC codes at the same code rate. As shown in Fig. \ref{fig::12864ldpc}, when $n=128$, the EW-GNN decoder still outperforms the BP decoder and the NBP decoder in terms of BER. Furthermore, when $n=256$ as illustrated in Fig. \ref{fig::256128ldpc}, EW-GNN outperforms BP and NBP by $0.3 ~\mathrm{dB}$ and $0.2 ~\mathrm{dB}$ of the coding gain at {high SNRs}, respectively. Therefore, we highlight that even though EW-GNN is trained with an extremely short code ($n=32$), {it can achieve better} BER performance when applied to longer codes {without fine-tuning}. 

\subsubsection{High-Scalability Performance}
{{As shown in Figs. 8-9,} the EW-GNN scheme exhibits significant scalability, allowing a decoder trained on one code to be used for another without retraining.}
{By leveraging the message-passing structure \cite{mpnn}, the EW-GNN decoder learns at nodes from the immediate local neighborhood information {and the} hidden parameters shared by all edges in the Tanner graph. 
{Thus,} the number of trainable parameters is independent of input graph size.}
{The EW-GNN decoder offers a versatile and efficient solution for decoding linear block codes of varying lengths, reducing the need for retraining and adapting to different code lengths.
The ability to generalize from short to long codes ensures that EW-GNN can maintain high performance across different coding scenarios, making them suitable for a wide range of applications in wireless communications.
}

\vspace{-0.6em}
\subsection{GNN-DRL Auto-encoder} \label{sec::ae_performance}
\vspace{-0.2em}

{W}e first evaluate the {performance} of the proposed auto-encoder framework by training it on a $(32,16)$ code {and compare it to the LDPC code.} 
{Next}, we extend the application of the auto-encoder model to a $(63, 45)$ code and compare its performance against a standard BCH code.
{Finally}, we analyze the impact of its {individual} components {of} encoder and decoder separately, and {elucidate how the} iterative training procedure {enhances the performance of auto-encoder}.

\subsubsection{$(32, 16)$ Codes}
\begin{figure}[t]
\centering
\includegraphics[width=0.39\textwidth]{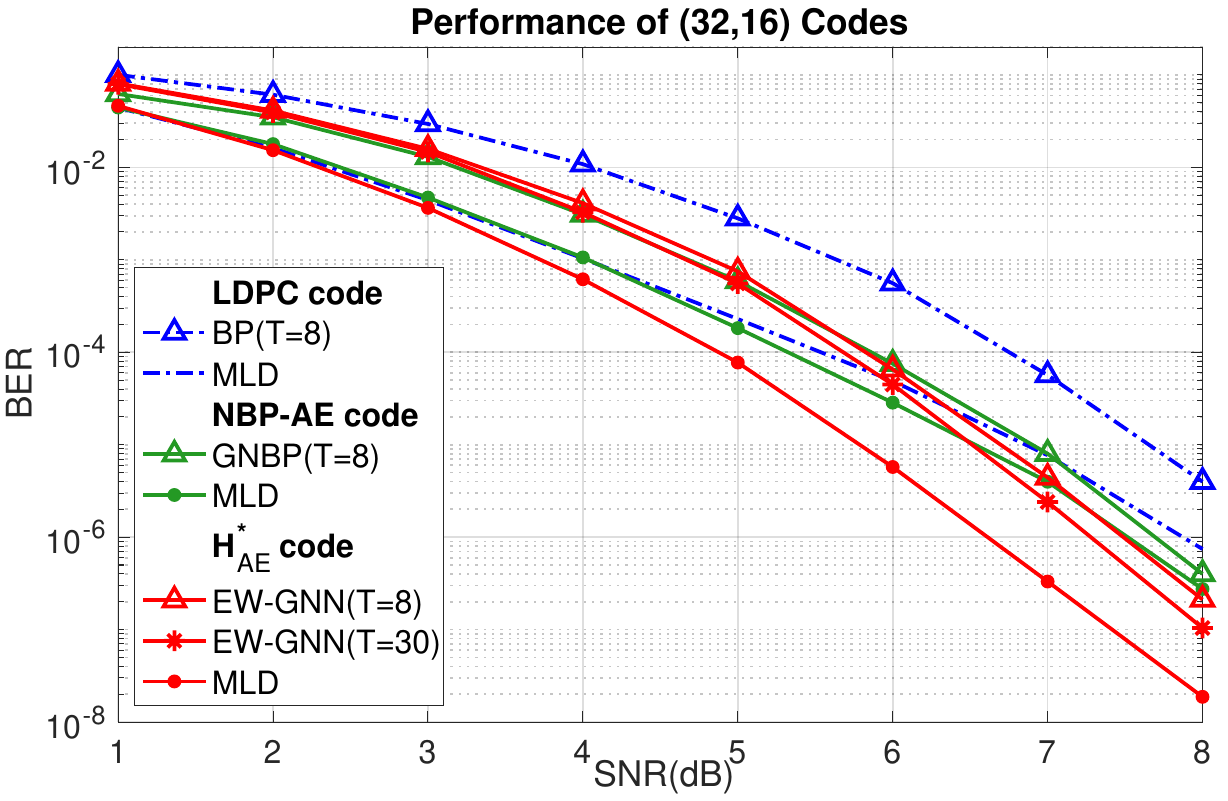}
\vspace{-0.4em}
\caption{BER performance for designed $(32, 16)$ linear block codes. {$\mathbf{H}^*_{\mathrm{AE}}$ code is generated by our auto-encoder scheme with three training iterations.}}
\vspace{-1em}
\label{fig::ae_3216}
\end{figure}
We first train an auto-encoder model with $\tau = 3$ training iterations {to produce $(32,16)$ codes}. 
Two benchmarks are considered {for comparison}: the {CCSDS} $(32,16)$ LDPC code {paired with} the BP decoder, and {the} NBP-AE \cite{nbp_ae2022}.
{NBP-AE is equipped with the} Gated Neural Belief Propagation (GNBP) decoder, which is an improved trainable version of BP. 
All these decoders, BP, GNBP, and our EW-GNN, {are} iterative message-passing {decoders implemented with 8 decoding iterations}.
{In the DRL-based code designer, the parity-check matrix of the CCSDS $(32,16)$ LDPC is utilized as the state initialization matrix.}
As depicted in Fig. \ref{fig::ae_3216}, our auto-encoder outperforms the traditional {LDPC} coding system by a {coding gain of $1~\mathrm{dB}$ under message-passing decoding.} 
Additionally, when compared with the NBP-AE system, our model shows a {coding gain} of approximately $0.21~\mathrm{dB}$ at high SNRs.
This coding gain can be attributed to the the reduced number of short cycles {in the Tanner graph of the code generated by our DRL-based code designer.}
{Specifically,} the code generated by our auto-encoder {after 3 training iterations}, denoted by $\mathbf{H}^*_\mathrm{AE}$, {has} only $77$ 4-step cycles, {which is }significantly fewer than $172$ short cycles in the code designed by NBP-AE.
Furthermore, {in} the proposed auto-encoder, we can directly apply the trained {EW-GNN} decoder to the designed code with a larger number of decoding iterations than that used in training.
{As shown in Fig. \ref{fig::enc_3216},} by {increasing the number of} EW-GNN {decoding iterations to 30}, we can achieve a $0.22~\mathrm{dB}$ {further} gain compared to {using} an 8-iteration {EW-GNN} decoder.

{We also test the MLD performance} of all these $(32,16)$ codes. Despite our auto-encoder model being trained to minimize the EW-GNN decoding BER, {generated codes also demonstrate} robust BER performance under MLD. 
Notably, our $\mathbf{H}^*_\mathrm{AE}$ code has coding gains of $1.28~\mathrm{dB}$ and $0.93~\mathrm{dB}$ over the CCSDS LDPC code and the $(32,16)$ NBP-AE code, respectively, at high SNRs.
{Moreover,}{ the proposed auto-encoder is capable of reconciling the decoding efficiency achieved by a suboptimal decoder (e.g., iterative decoding) with the decoding performance achieved by an optimal decoder (e.g., MLD).}
{For example,} when using the EW-GNN decoder with $T=30$ iterations for $\mathbf{H}^*_\mathrm{AE}$, it achieves a BER that is superior to the MLD performance of LDPC for SNRs above $6~\mathrm{dB}$, with a coding gain of up to $0.63~\mathrm{dB}$.

\subsubsection{$(63, 45)$ Codes}

\begin{figure}[t]
\centering
\includegraphics[width=0.39\textwidth]{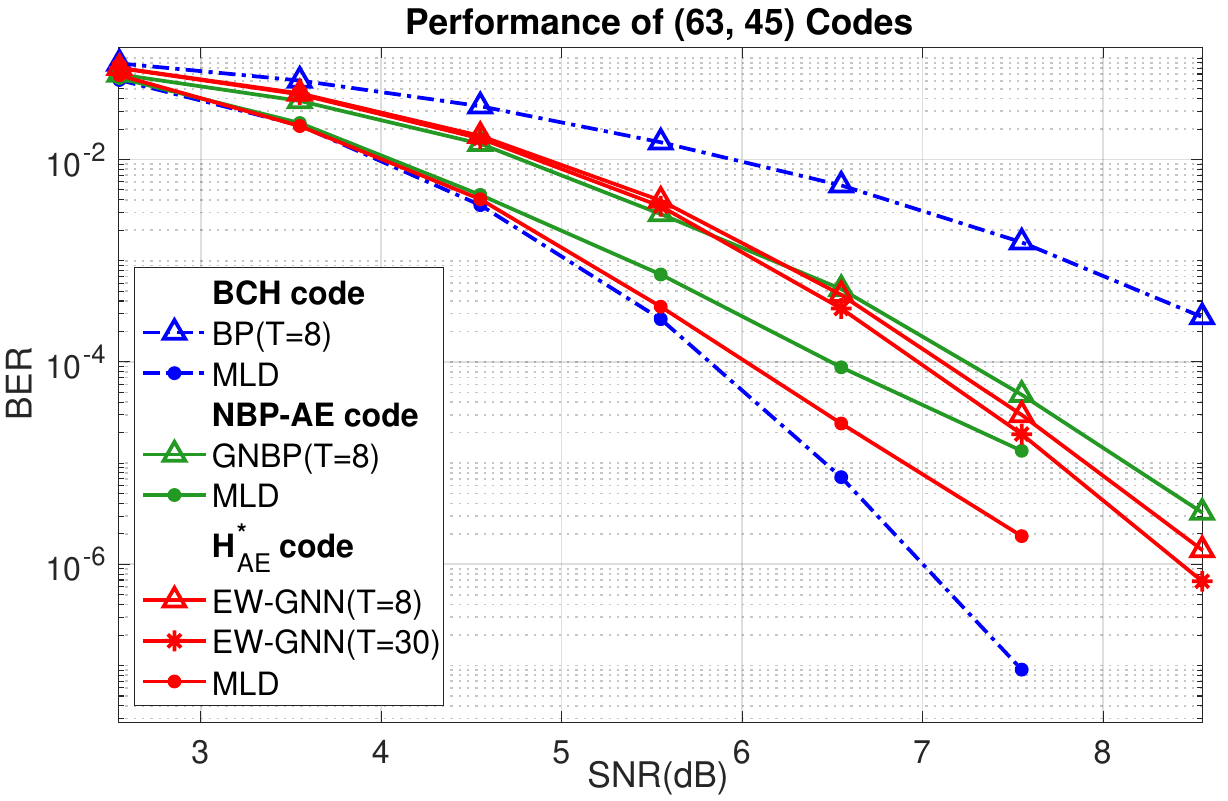}
\vspace{-0.6em}
\caption{BER performance for designed $(63, 45)$ linear block codes. {$\mathbf{H}^*_{\mathrm{AE}}$ code is generated by our auto-encoder scheme with three training iterations.}}
\vspace{-1em}
\label{fig::ae_6345}
\end{figure}

{W}e also train an auto-encoder model {using} the parity-check matrix of the $(63,45)$ BCH \cite{rptudataset} as the state initialization.
As depicted in Fig. \ref{fig::ae_6345}, the proposed auto-encoder, {i.e., the} learned $(63,45) ~ \mathbf{H}^*_\mathrm{AE}$ with its associated EW-GNN decoder, is compared with {the} $(63,45)$ BCH {paired with} BP and {the} $(63,45)$ NBP-AE system \cite{nbp_ae2022}.
Our auto-encoder scheme significantly outperforms the standard BCH with BP and the neural NBP-AE system, achieving coding gains of up to $1.81~\mathrm{dB}$ and $0.28~\mathrm{dB}$ at high SNRs, respectively.

{Figure \ref{fig::ae_6345} presents} the BER {of MLD} for all these $(63,45)$ codes.
However, unlike the {case} of $(32, 16)$ codes, the MLD performance of the learned code is worse than that of the BCH code. 
This {is reasonable because the BCH code is the best-known linear code {with} the highest MHD compared with all other existing codes.}
{Under MLD}, the BCH code {can achieve the} normal approximation {bound} \cite{shirvanimoghaddam2018short} {at this short block length}.
In addition, our auto-encoder model is specifically trained to optimize coding performance with the EW-GNN decoder, {rather than} the MLD.
The learning process tends to generate matrices with lower density and fewer short cycles, which is ideal for BP and EW-GNN to achieve an acceptable performance-complexity trade-off. 

\subsubsection{{Impact of each Components}}
\begin{figure}[t]
\centering
\includegraphics[width=0.39\textwidth]{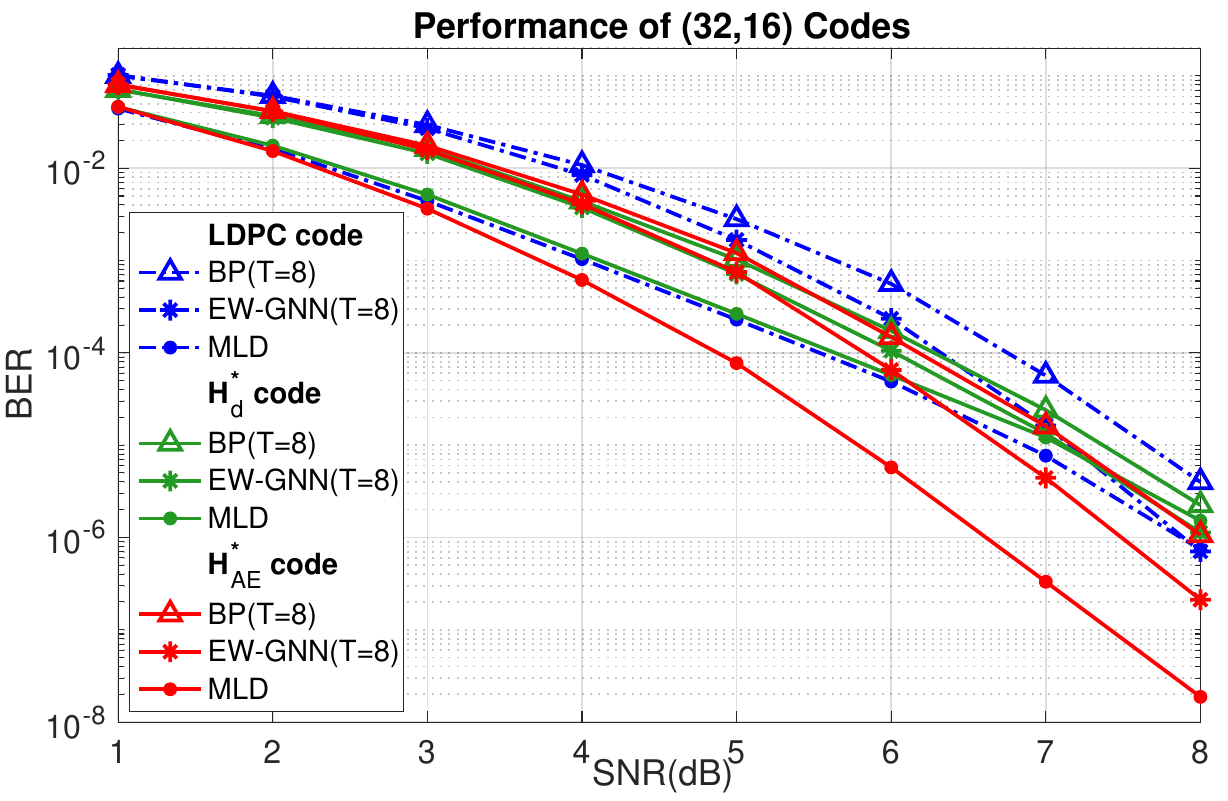}
\vspace{-0.6em}
\caption{Evaluate the relative contributions of the DRL-based code designer and the EW-GNN decoder.}
\vspace{-1em}
\label{fig::ae_elements_study}
\end{figure}

\begin{figure}[t]
\centering
\includegraphics[width=0.39\textwidth]{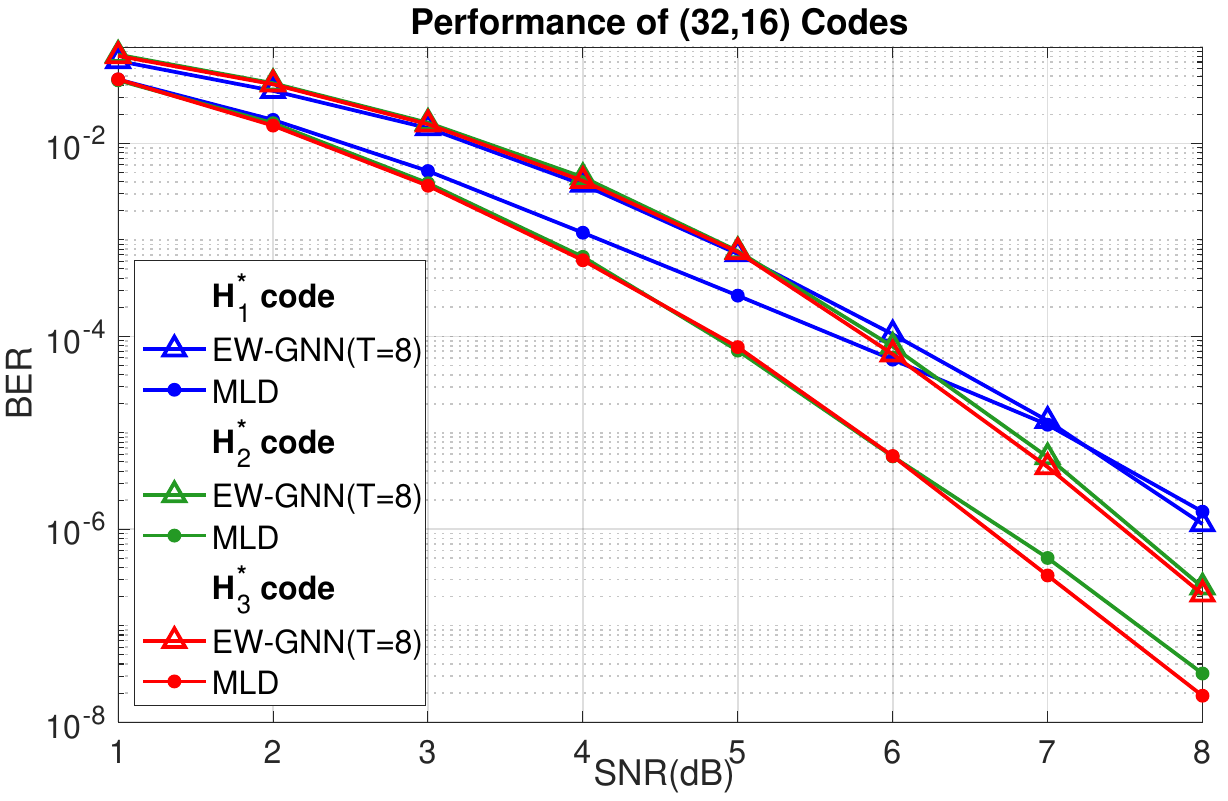}
\vspace{-0.6em}
\caption{Evaluate the relative contributions of the iterative training method.}
\vspace{-1.2em}
\label{fig::ae_iter_study}
\end{figure}
In Fig. \ref{fig::ae_elements_study}, we analyze the impact of various elements {over the performance of the} auto-encoder by comparing {different code-decoder pairs of $(32,16)$}. 
In the figure, $\mathbf{H}^*_{\mathrm{AE}}$ is the code produced by our auto-encoder scheme with three training iterations, while $\mathbf{H}^*_d$ is produced by the DRL-based code designer using BP-decoding BER reward in one-shot training. When applied to $\mathbf{H}^*_{\mathrm{AE}}$, the EW-GNN decoder is iteratively trained in the auto-encoder, while when applied to $\mathbf{H}^*_{d}$, the EW-GNN decoder is fine-tuned for $\mathbf{H}^*_{d}$. 
    
    
    
    
    


{All the schemes using the code designed from DRL, including $\mathbf{H}^*_{\mathrm{AE}}$ and $\mathbf{H}^*_{d}$, can achieve better performance than the traditional LDPC with BP.}
{O}ur auto-encoder model {(i.e., $\mathbf{H}^*_\mathrm{AE}$ with EW-GNN)} achieves the best BER performance {among all schemes}.
In addition, {the} code generated by our auto-encoder framework, $\mathbf{H}^*_\mathrm{AE}$, has better BER results than its counterparts, under all three different decoders. 
Especially, the BER result of the $\mathbf{H}^*_\mathrm{AE}$ under MLD {significantly} outperforms that of all other coding schemes with substantial coding gains.
{Furthermore, as shown in Fig. \ref{fig::ae_elements_study}, when replacing the LDPC code with the learned code while still decoding with BP, the BER performance improves primarily at low SNRs. Nevertheless, when comparing EW-GNN to BP in decoding LDPC code, the performance gain is significant only at high SNRs.}
{In contrast,} the auto-encoder model can have a superior BER performance under a wide range of SNRs.  
Therefore, both the trainable code {and the} trainable EW-GNN contribute to the final {outstanding} performance of the auto-encoder.

To evaluate {the impact of the iterative training procedure}, we compare the performances of various auto-encoder models with different numbers of training iterations {as demonstrated in Fig.} \ref{fig::ae}.
{As shown {in} Fig. \ref{fig::ae_iter_study}}, $\mathbf{H}^*_{2}$ significantly outperforms $\mathbf{H}^*_{1}$ under both EW-GNN decoder and MLD.
{However,} the coding gain between $\mathbf{H}^*_{2}$ and $\mathbf{H}^*_{3}$ diminishes, {indicating that the performance of both code and decoder has nearly converged to the optimal point.} 
{This fast convergence} can be attributed to the high scalability of the EW-GNN decoder.
As {discussed} in Section \ref{sec::dec_simulation}, EW-GNN exhibits high scalability across various codes. 
Consequently, different codes as training inputs do not significantly affect {the optimality of the EW-GNN decoder}.
Thus, from the {second} training iteration onwards, performance gain {is} primarily contributed by {the DRL-based code designer.} 

\vspace{-0.6em}
\section{Conclusion} \label{sec::conclusion}
\vspace{-0.2em}
In this paper, we proposed an innovative auto-encoder framework that leverages GNN and DRL to develop {high-performance} channel coding schemes for short linear block codes. 
First, we introduced a DRL-based code designer, where a GNN agent generates the binary parity-check matrix by iteratively flipping its elements. 
Simulation results demonstrated that the DRL-designed short code exhibit superior BER performance {with a $0.83~\mathrm{dB}$ coding gain} compared to the standard LDPC code under MLD.
Furthermore, we developed a scalable {EW-GNN decoder} for short linear block codes. Simulation results show that EW-GNN can improve BER performance compared to the conventional BP and the NBP decoder for short BCH and LDPC codes. 
{For example, when decoding $(63,51)$ BCH code, the proposed EW-GNN increases the coding gain by 1.2 dB than BP.}
{A well-trained EW-GNN can decode longer or more complex codes without retraining, enabling its application in practical communication systems with dynamic code lengths.}
Finally, the GNN-DRL auto-encoder integrates the DRL-based code designer and the EW-GNN decoder through an iterative joint training process, ensuring simultaneous optimization of both the encoder and decoder. 
{Simulations show that our proposed} auto-encoder framework {achieves} better BER results than multiple traditional coding systems {and the NBP-AE} {from the literature, while keeping the low decoding complexity similar to that of BP.}
{Particularly, our} auto-encoder system can achieve a coding gain of $0.63~\mathrm{dB}$ compared to the standard {$(32,16)$} LDPC with MLD.

\vspace{-1.2em}
\bibliographystyle{IEEEtran}
\bibliography{ref}

\end{document}